\documentclass[11pt]{article}
\usepackage[margin=1in]{geometry}

\usepackage{amsmath}
\usepackage{amssymb}
\usepackage[english]{babel}
\usepackage{bm}
\usepackage{caption}
\usepackage{ dsfont }
\usepackage{comment}
\usepackage{enumerate}
\usepackage{graphicx,comment}
\usepackage{graphicx}
\usepackage{hyperref}
\usepackage{lipsum}
\usepackage{longtable,tabularx}
\usepackage{ mathrsfs }
\usepackage[version=4]{mhchem}
\usepackage{multirow}
\usepackage{multimedia}
\usepackage{rotating}
\usepackage{siunitx}
\usepackage{soul}
\usepackage{supertabular,booktabs}
\usepackage{tikz}
\usepackage{subcaption}
\usepackage{authblk}   
\usepackage{titling}   

\usepackage{booktabs}
\usepackage{multirow}
\usepackage{makecell}

\DeclareSymbolFont{CMMforNu}{OML}{cmm}{m}{it}
\SetSymbolFont{CMMforNu}{bold}{OML}{cmm}{b}{it}
\DeclareMathSymbol{\nuCM}{\mathord}{CMMforNu}{'027}
\usepackage[outline]{contour}
\def\nu{\mathord{\contourlength{0.0043em}\contour{black}{$\nuCM$}}}

\usepackage{authblk}   

\newcommand{\deriv}[2]{\frac{\mathrm{d} #1}{\mathrm{d} #2}}

\usepackage[utf8]{inputenc}
\usepackage[T1]{fontenc}
\usepackage{mathptmx}
\usepackage{etoolbox}

\usepackage{algorithm}
\usepackage{algpseudocode}

\title{\Large\bfseries Trajectory-Optimized Time Reparameterization for Learning-Compatible Reduced-Order Modeling of Stiff Dynamical Systems}

\author{
  Joe Standridge\\
  Texas A\&M University, College Station, Texas 77840\\
  \and
  Daniel Livescu\\
  Los Alamos National Laboratory, Los Alamos, New Mexico 87545\\
  \and
  Paul G. A. Cizmas\\
  Texas A\&M University, College Station, Texas 77840
}

\date{\today}

\begin{document}
\maketitle

\begin{abstract}
Stiff dynamical systems present a fundamental challenge for machine-learning reduced-order models (ML-ROMs), as explicit time integration becomes unstable in stiff regimes while implicit integration within learning loops is computationally expensive and often degrades training efficiency. Time reparameterization (TR) offers a promising alternative by transforming the independent variable so that rapid physical-time transients are spread over a stretched-time coordinate, enabling stable explicit integration on uniformly sampled grids. Although several TR strategies have been proposed, their effect on learnability in neural ODE–based ROMs remains incompletely understood. This work systematically investigates time reparameterization as a stiffness-mitigation mechanism for neural ODE reduced-order modeling and introduces a trajectory-optimized TR (TOTR) formulation. The proposed approach casts time reparameterization as an optimization problem posed in arc-length coordinates, in which a traversal-speed profile is selected to directly penalize acceleration in stretched time. By targeting the smoothness of the dynamics actually integrated during training, this formulation produces time maps and reparameterized trajectories that are better conditioned and easier to learn than other TR proposed methods. TOTR is evaluated on three representative stiff benchmark problems: a parameterized stiff linear system, the van der Pol oscillator, and the HIRES chemical kinetics model. Across all cases, the proposed approach consistently yields smoother reparameterizations and improved physical-time predictions under identical training regimens than other recent TR approaches. Quantitative results demonstrate loss reductions of one to two orders of magnitude compared to benchmark algorithms, particularly in regimes with severe or monotone stiffness. These results highlight that effective stiffness mitigation in ML-ROMs depends critically on the regularity and learnability of the time map itself, and that optimization-based TR provides a robust framework for explicit reduced-order modeling of multiscale dynamical systems.
\end{abstract}


\section{Introduction}
Stiff dynamical systems arise throughout science and engineering, from chemical kinetics and incompressible flows to control, and remain a core obstacle in high-fidelity simulation due to widely separated time scales. Rapid transients alongside slow modes can make accurate simulation computationally prohibitive, motivating reduced-order models (ROMs) that remain efficient and numerically stable in stiff regimes.

Stiffness has long been tied to stability-driven step-size restrictions: Dahlquist’s foundational work formalized absolute stability and showed why explicit methods can become impractical for stiff problems~\cite{Dahlquist1961}. Subsequent developments emphasized identifying stiff regimes and using stable integration strategies to obtain tractable simulations~\cite{petzold1983,HairerWannerII,AscherPetzold}. Later analyses clarified that stiffness is diagnosed by excessive computational cost and step-size restriction rather than purely spectral criteria~\cite{UsersViewStiffODEs,Soderlind2015}, and that nonlinearity can create numerical difficulty even when classical indicators appear mild~\cite{HighamTrefethen1993}. Modern perspectives therefore view stiffness as a manifestation of multiscale dynamics and fast transients that constrain admissible time steps and solver design~\cite{hairer2013}.

These challenges persist in ROMs: projection-based methods often inherit full-order stiffness, limiting speedups unless additional structure is exploited~\cite{parametricROM}. In chemical kinetics, stiffness is commonly addressed through explicit time-scale separation (e.g., quasi-steady-state, partial equilibrium, slow manifolds)~\cite{MaasPope,pope04,Law}, but such approximations are not generally available. Recently, neural ordinary differential equations (NODEs)~\cite{Chen2018} have emerged as a flexible data-driven ROM paradigm, yet stiffness can dominate both training and inference, motivating stiffness-aware NODE/ROM design.

Recent NODE-ROM work addresses stiffness through several strategies. Some approaches modify training to improve numerical stability, e.g., scaling and stabilized gradient computations for stiff benchmarks~\cite{stiffNode}. Others incorporate implicit or stiff-stable solvers into the learning loop, including implicit single-step schemes and stiff-stable exponential-integration variants~\cite{FronkPetzoldImplicit,FronkPetzoldExp}. A third direction seeks representations in which the learned dynamics are effectively less stiff, for example via autoencoder-based latent coordinates coupled to NODE surrogates~\cite{Vijayarangan2025AE}.

Caldana and Hesthaven~\cite{caldana2025} showed that time reparameterization (TR)—where the dynamics are transformed to a new time coordinate—can substantially reduce stiffness for NODE-based ROMs. Related ideas have also been applied in Gaussian process surrogates for stiff systems by Cotr\'es Garc\'ia et al.~\cite{garcia2025}, suggesting that stiffness can sometimes be mitigated by transforming the independent variable.

Time reparameterization has deep classical roots. In celestial mechanics, Sundman-type transformations rescale time to regularize singular dynamics in the $N$-body problem~\cite{Sundman1912}. More broadly, reparameterizing time is a standard analytical and geometric tool for studying trajectories and asymptotics without changing phase-space paths~\cite{Arnold1989}, and stretched/logarithmic time scalings are common in multiscale and singular perturbation analysis~\cite{KevorkianCole}. Historically, however, using such transformations in reduced or data-driven modeling was hindered by the need to invert time maps for off-reference data.

Within the neural ODE framework, TR becomes practical: transformed-time trajectories can be learned and deployed directly, avoiding explicit inversion while improving numerical tractability. This matters because typical neural ODE pipelines favor explicit integration and uniformly sampled data for efficient batching and gradient computation~\cite{Chen2018,rubanova2019,Rackauckas2021}, whereas stiff full-order dynamics often require implicit/adaptive solvers, producing nonuniform sampling and expensive gradients~\cite{ANODE2019,stiffNode}. By representing stiff trajectories in a transformed coordinate $\tau$, TR can enable uniform sampling and stable explicit integration, restoring conditions favorable for efficient training and mitigating stiffness in the learned dynamics.

Despite promising results, TR for learning-based ROMs remains only partially understood: existing approaches show that suitable time maps can reduce stiffness and improve learnability, but the formulation, optimization, and deployment of time transformations—and their impact on fidelity for full-order dynamics—have not been systematically studied. This gap motivates the present work. We investigate time reparameterization for mitigating stiffness in neural ODE–based ROMs and propose a formulation that treats stiffness in transformed time as an explicit optimization objective. We evaluate multiple TR strategies across representative stiff dynamical systems.

This paper introduces the TR formulation and its effect on stiffness in Section~\ref{sec:tr_formulation}, along with two benchmark TR strategies from the literature. The proposed algorithm is described in Section~\ref{sec:method}. Section~\ref{sec:experimental} details training procedures and evaluation metrics, and Section~\ref{sec:results} presents numerical results and comparisons. Implications and limitations are discussed in Section~\ref{sec:discussion}, with conclusions in Section~\ref{sec:conclusion}.


\section{Background}
\label{sec:tr_formulation}
\subsection{Neural ODE setting}
\label{subsec:node_setting}

Consider the full-order dynamical system:
\begin{equation}
  \deriv{\bm{y}}{t} = \bm{f}(\bm{y}, t; \mu), 
  \qquad \bm{y}(0;\mu)=\bm{y}_0(\mu),
  \qquad \bm{y}(t;\mu)\in\mathbb{R}^d ,
  \label{eq:fom}
\end{equation}
where $\mu$ denotes a set of parameters. In the neural ODE framework, the vector
field $\bm{f}$ is approximated by a neural network $\bm{f}_{NN}$ and learned from
trajectory data by minimizing a loss defined on solution observations. A reduced order model is obtained:
\begin{equation}
  \deriv{\bm{z}}{t} = \bm{f}_{NN}(\bm{z}, t; \mu), 
  \qquad \bm{z}(0;\mu)=\bm{y}_0(\mu),
  \qquad \bm{z}(t;\mu)\in\mathbb{R}^d ,
  \label{eq:rom}
\end{equation}
such that $\bm{z}\approx\bm{y}$ after integration. Training and prediction of \eqref{eq:rom} are performed by integrating the resulting ODE forward in time. For stiff systems, however, direct application of the NODE framework is challenging.
Rapid transients in physical time force explicit integrators to take extremely small time steps for stability, while implicit or adaptive solvers introduce nonuniform time grids and complicate gradient computation. These difficulties motivate modifications to the standard NODE formulation that address stiffness at the level of the dynamical representation.

\subsection{Time reparameterization}
\label{subsec:time_reparam}

A strictly increasing map between physical time $t$ and a stretched time
coordinate $\tau$ is introduced:
\begin{equation}
  t = t(\tau;\mu), 
  \qquad \frac{\mathrm{d}t}{\mathrm{d}\tau} = \alpha(\bm{y},\tau, \mu) > 0,
  \qquad t(0;\mu)=0,
  \label{eq:time_map}
\end{equation}
where $\alpha(\tau,\bm{y};\mu)$ is a time dilation function. Defining the reparameterized trajectory as:
\begin{equation}
  \tilde{\bm{y}}(\tau;\mu) := \bm{y}(t(\tau;\mu);\mu),
  \label{eq:y_def}
\end{equation}
the chain rule gives
\begin{equation}
  \frac{\mathrm{d}\tilde{\bm{y}}}{\mathrm{d}\tau}
   = \alpha(\bm{\tilde{y}},\tau;\mu)\,\bm{f}(\tilde{\bm{y}}, t(\tau;\mu); \mu).
  \label{eq:chain_rule_general}
\end{equation}

Equivalently, the reparameterized dynamics may be written as the augmented system:
\begin{subequations}
\begin{align}
  \frac{\mathrm{d}\tilde{\bm{y}}}{\mathrm{d}\tau} &=\alpha(\bm{\tilde{y}},\tau;\mu)\bm{f}(\tilde{\bm{y}}, t(\tau;\mu); \mu), 
  \qquad \tilde{\bm{y}}(0)=\bm{y}_0(\mu), \label{eq:aug_u}\\
  \frac{\mathrm{d}t}{\mathrm{d}\tau} &=\alpha(\bm{\tilde{y}},\tau;\mu),
  \qquad t(0)=0. \label{eq:aug_t}
\end{align}
\label{eq:augmented_system}
\end{subequations}
Since $\alpha(\tau,\bm{y};\mu)>0$, the mapping $t(\tau;\mu)$ is invertible and the original physical-time solution is recovered via $\bm{y}(t)=\tilde{\bm{y}}(\tau(t))$.

At an intuitive level, stiffness corresponds to rapid changes occurring over short
physical-time intervals, which impose severe step-size restrictions on explicit integrators. Time reparameterization mitigates this effect by redistributing temporal resolution: choosing $\alpha$ to be small in stiff regions forces physical time to advance slowly with respect to $\tau$, spreading sharp transients over a larger $\tau$-interval. Conversely, in slowly varying regions, larger values of $\alpha$ permit faster advancement of physical time. In doing so, a proper choice of $\alpha$ can transform a stiff trajectory into a one with more uniform time scales in the stretched time coordinate.  This decreases the time scale disparity and, by extension, the stiffness present.

From a stability perspective, $\alpha$ scales the entire right-hand side of \eqref{eq:aug_u}, and therefore scales the eigenvalues governing explicit stability. Selecting $\alpha\ll 1$ in stiff regions reduces the magnitude of the effective eigenvalues in $\tau$, enlarging the admissible step size for explicit integration in the reparameterized system. This scaling viewpoint motivates the design of $\alpha$ to ``flatten'' disparate time scales and reduce stiffness in the transformed dynamics. 

Within the neural ODE framework, time reparameterization may be incorporated directly into the learned dynamical system. Specifically, both the rescaled vector field $\bm{f}$ and the time dilation function $\alpha$ are approximated using neural networks. Let $\bm{f}_{\mathrm{NN}}(\tilde{\bm{z}}, \tau;\mu)$ denote a neural network approximation of the stretched-time vector field, and let $ \alpha_{\mathrm{NN}}(\tilde{\bm{z}},\tau;\mu)$ denote a neural network  parameterization of the time dilation function, constrained to be strictly positive.

Substituting these approximations into the reparameterized dynamics yields the neural ODE in stretched time:
\begin{subequations}
\begin{align}
  \frac{d\tilde{\bm{z}}}{d\tau}
  &= \bm{f}_{\mathrm{NN}}(\tilde{\bm{z}}, \tau;\mu),
  \label{eq:node_tr_y}\\
  \frac{dt}{d\tau}
  &= \alpha_{\mathrm{NN}}(\tilde{\bm{z}},\tau;\mu),
  \label{eq:node_tr_t}
\end{align}
\label{eq:node_tr_system}
\end{subequations}
with initial conditions $\tilde{\bm{z}}(0)=\bm{y}_0(\mu)$ and $t(0)=0$.

The neural network $\bm{f}_{\mathrm{NN}}$ is trained to represent the system dynamics in the stretched-time coordinate, while $\alpha_{\mathrm{NN}}$ learns a time dilation that reconstructs the physical-time evolution induced by the chosen reparameterization strategy. To guarantee positivity of the dilation function and preserve monotonicity of the time map, the output of $\alpha_{\mathrm{NN}}$ is passed through an exponential activation.

Training proceeds by integrating the augmented system \eqref{eq:node_tr_system} forward in $\tau$ and minimizing a loss defined on observations of $\tilde{\bm{z}}(\tau;\mu)$  and $t(\tau;\mu)$. This formulation provides a neural ODE framework in which stiffness is addressed at the level of trajectory parametrization, enabling explicit integration and uniform sampling in $\tau$ while preserving fidelity to the original physical-time dynamics.

\subsection{Solver-directed time reparameterization}

In practice, $\alpha$ is not prescribed \textit{a priori}. Instead, it can be inferred from full-order simulations by exploiting information already produced by stiff integrators.

Caldana and Hesthaven~\cite{caldana2025} construct $\tau$ from the adaptive time grid generated by a stiff-stable implicit solver applied to the full-order system. Let
\[
  0=t_0 < t_1 < \cdots < t_N = T
\]
be the adaptive grid produced by the implicit integrator. A normalized stretched time is
then defined by
\begin{equation}
  \tau_n := \frac{n}{N}\,\tau_f, \qquad t(\tau_n)=t_n,\qquad n=0,\dots,N,
  \label{eq:uniform_tau_grid}
\end{equation}
where $\tau_f$ is the chosen stretched final time. The full-order trajectory is subsequently re-sampled on an uniform $\tau$ grid.
The implied time-velocity $\alpha(\bm{\tilde{y}},\tau;\mu)=\mathrm{d}t/\mathrm{d}\tau$ is then learned as a function of state so that physical time can be recovered during deployment. The algorithm is summarized in Algorithm~\ref{alg:caldana}.

\begin{algorithm}[h!]
\caption{Solver-Directed Time Reparameterization}
\label{alg:caldana}
\begin{algorithmic}[1]
\Require Physical-time trajectory $\{(t_n,\bm{y}_n)\}_{n=0}^N$, target stretched final time $\tau_f$
\Ensure Reparameterized trajectory $\tilde{\bm{y}}(\tau)$ and time map $t(\tau)$ on a uniform $\tau$ grid

\State Compute $\tau$ at each point $\tau_n := \frac{n}{N}\,\tau_f$, such that $t(\tau_n)=t_n$ and $\tilde{\bm{y}}(\tau_n)=\bm{y}_n$.
\State Interpolate $\tilde{\bm{y}}(\tau)$  and $t(\tau)$ onto uniform grid $\tau\in[0,\tau_f]$.
\end{algorithmic}
\end{algorithm}
\subsection{Extrema-based time reparameterization}
An alternative class of time reparameterization strategies is based on geometric properties of the solution trajectory. Given a physical-time trajectory $\bm{y}(t)$, one may define its arc length:
\begin{equation}
  s(t) = \int_{0}^{t} \sqrt{1+\left\| \deriv{\bm{y}}{t'} \right\|^2 \,} \mathrm{d} t',
  \label{eq:arc_length}
\end{equation}
which measures the cumulative variation of the state along the trajectory. Both $\bm{y}$ and $t$ may be subject to nondimensionalzation prior to computing arc length to ensure each ODE contributes roughly equally to the arc length.  Reparameterizing the solution by arc length,
\begin{equation}
  \bar{\bm{y}}(s) := \bm{y}(t(s)),
  \label{eq:arc_param}
\end{equation}
naturally redistributes resolution toward regions of rapid state variation, yielding a trajectory that is more uniformly resolved with respect to $s$. Arc-length coordinates therefore capture local geometric features of the dynamics and reduce nonuniformity induced by disparate physical time scales.

Cort\'es Garc\'ia et al.~\cite{garcia2025} observe that although the arc-length–parameterized trajectory $\bar{\bm{y}}(s)$ is smooth, stiffness can persist near the extrema of $\bm{y}$. In these regions, the derivative $\mathrm{d}\bar{\bm{y}}/\mathrm{d}s$ may exhibit rapid variations relative to the rest of the trajectory. To mitigate this effect, the authors propose selecting the reparameterization such that:
\begin{equation}
  \deriv{\tau}{s}\Big| _{s_{\mathrm{ext}}} = 0,
  \label{eq:root_condition}
\end{equation}
at arc-length locations $s_{\mathrm{ext}}$ corresponding to the $N$ extrema of $\bm{y}$ such that

\begin{align}
\begin{alignedat}{5}
  \tau(0)   &= 0,        &\qquad \tau(s_1)   &= \frac{s_1}{S}\tau_f, 
  &\quad \ldots \hspace{.2em} , \quad \tau(s_N)   &= \frac{s_N}{S}\tau_f, 
  &\qquad \tau(S)   &= \tau_f, \\
  \tau'(0)  &= 0,        &\qquad \tau'(s_1)  &= 0, 
  &\quad \ldots \hspace{.2em}, \quad \tau'(s_N)  &= 0, 
  &\qquad \tau'(S)  &= 0 .
\end{alignedat}
\end{align}

    In practice, these conditions are enforced by constructing cubic or quintic splines between $s_n$ and $s_{n+1}$. Such splines preserves monotonicity while spreading stiff regions over a larger $\tau$-interval. The two reparameterizations $t(s)$ and $\tau(s)$ can be composed to obtain $t(\tau)$. The algorithm is summarized in Algorithm~\ref{alg:garcia}.

\begin{algorithm}[h!]
\caption{Extrema-Based Time Reparameterization}
\label{alg:garcia}
\begin{algorithmic}[1]
\Require Physical-time trajectory $\{(t_n,\bm{y}_n)\}_{n=0}^N$, target stretched final time $\tau_f$
\Ensure Reparameterized trajectory $\tilde{\bm{y}}(\tau)$ and time map $t(\tau)$ on a uniform $\tau$ grid

\State Compute discrete arc-length samples $\{s_n\}$ from $\bm{y}$ and construct the arc-length trajectory $\bar{\bm{y}}(s)$.
\State Interpolate $\bar{\bm{y}}(s)$ onto a uniform arc-length grid $s\in[0,S]$.

\State Identify the extrema of $\bar{\bm{y}}(s)$.
\State Construct $\tau$ using cubic or quintic splines.
\State Invert $\tau(s)$ to find $s(\tau)$.
\State Construct the reparameterized trajectory $\tilde{\bm{y}}(\tau)=\bar{\bm{y}}(s(\tau))$ and the physical-time map $t(\tau)=t(s(\tau))$ by composing the two reparameterizations.
\State Interpolate $\tilde{\bm{y}}(\tau)$  and $t(\tau)$ onto uniform grid $\tau\in[0,\tau_f]$.
\end{algorithmic}
\end{algorithm}

\section{Method}
\label{sec:method}\subsection{Proposed time reparameterization algorithm}
Existing time reparameterization strategies reduce stiffness indirectly by redistributing temporal resolution along the trajectory. This redistribution is determined  according to solver behavior or geometric heuristics. While these approaches are effective in many settings, they do not explicitly control the smoothness of the reparameterized trajectory with respect to the new independent variable. 

Consequently, solver-driven reparameterization may become sensitive to numerical noise or parametric variation in certain regimes. Purely heuristic approaches can also fail for important cases. For example, extrema-based methods may fail to detect or adequately resolve isolated stiff events that are not associated with extrema.

In contrast, we propose a novel approach that treats time reparameterization itself as an optimization problem, which we call Trajectory Optimized Time Reparameterization (TOTR). We optimize $t(\tau)$ posed directly on a fixed $\bm{y}(t)$ such that the trajectory $ \tilde{\bm{y}}(\tau)$ becomes as smooth as possible in $\tau$. The central observation motivating this formulation is that stiffness appears as rapid variations in velocity and curvature with respect to the independent variable. Minimizing stiffness the stiffness of $ \tilde{\bm{y}}(\tau)$ in stretched time $\tau$ therefore amounts to constructing a reparameterization for which the transformed trajectory $\tilde{\bm{y}}(\tau)$ evolves as smoothly as possible.

To formalize this idea, we introduce the augmented state $\tilde{\bm{\varphi}}$
\begin{equation}
    \tilde{\bm{\varphi}}:=[y_1,y_2,...,y_n,t]^\intercal,
\end{equation}
where the time variable $t$ is treated as an additional component. The time map $t(\tau)$ can then be interpreted as an ordinary differential equation within this augmented system.

When seeking to maximize the smoothness of the trajectory, a natural objective is to minimize the squared acceleration of the reparameterized trajectory
\begin{equation}
  \mathcal{J}[\tau]
  =
  \int_{0}^{\tau_f}
  \left\|
    \frac{d^2 \tilde{\bm{\varphi}}}{d\tau^2}
  \right\|^2
  \, \mathrm{d}\tau.
  \label{eq:naive_stiffness}
\end{equation}
In doing so, rapid variations in the dynamics actually integrated in $\tau$ are directly penalized. In this formulation, the acceleration of the time map is also considered alongside the nominal state variables.

However, direct discretizing \eqref{eq:naive_stiffness} in stretched time leads to severe numerical difficulties. A  central-difference approximation of the second derivative introduces factors of $(\Delta\tau)^{-2}$, and squaring the expression produces $(\Delta\tau)^{-4}$. The resulting optimization problem therefore becomes severely ill-conditioned, particularly in stiff regions where small $\Delta\tau$ values are unavoidable. This leads to unstable gradients and poor numerical behavior.

To avoid these difficulties, the proposed method reformulates the objective \eqref{eq:naive_stiffness} in the arc-length domain $s$. Before performing the arc-length reparameterization, each component of $\tilde{\bm{\varphi}}$ may be nondimensionalized so that all components contribute roughly equally to the arc length. In practice, assigning equal weight to the state and time components of $\tilde{\bm{\varphi}}$ was found to distribute stretched-time resolution more uniformly across the trajectory.

Let $s\in[0,S]$ denote the arc-length coordinate along the trajectory, and let $\bar{\bm{\varphi}}(s)$ denote the concomitant arc-length parameterization. Rather than optimizing the time map $t(\tau)$ directly, we introduce a positive traversal speed profile:
\begin{equation}
  v(s) := \deriv{s}{\tau},
\end{equation}
which fully characterizes the time reparameterization. The arrival-time constraint becomes
\begin{equation}
  \tau_f = \int_0^S \frac{\mathrm{d}s}{v(s)}.
\end{equation}

Using a Frenet--Serret decomposition of the arc-length trajectory, the acceleration of $\bar{\bm{\varphi}}$ with respect to $\tau$ admits the representation
\[
\frac{\mathrm{d}^2\bar{\bm{\varphi}}}{\mathrm{d}\tau^2}
=
v\,v_s\,\mathbf{T}
+
\kappa(s)\,v^2\,\mathbf{N},
\]
where $\mathbf{T}$ and $\mathbf{N}$ are the unit tangent and principal normal vectors, $v_s=\mathrm{d}v/\mathrm{d}s$, and $\kappa(s)$ denotes the curvature of the trajectory in state space. Consequently,
\[
\left\|
\frac{\mathrm{d}^2\bar{\bm{y}}}{\mathrm{d}\tau^2}
\right\|^2
=
\left(v\,v_s\right)^2
+
\left(\kappa(s)\,v^2\right)^2.
\]

Substituting $\mathrm{d}\tau = \mathrm{d}s / v(s)$, the optimization problem can be written entirely in the arc-length coordinate as
\begin{equation}
  \mathcal{J}_s[v]
  =
  \int_0^S
  \left(
    v(s)v_s(s)^2
    +
    \kappa(s)^2\,v(s)^3
  \right)\,\mathrm{d}s,
  \quad
  \text{subject to }
  \int_0^S \frac{\mathrm{d}s}{v(s)}=\tau_f.
  \label{eq:Js_final}
\end{equation}

The two terms in \eqref{eq:Js_final} admit clear physical interpretations. The first penalizes rapid variations in traversal speed along the curve, corresponding to tangential acceleration, while the second penalizes acceleration induced by geometric curvature through the normal component. By construction, minimizing \eqref{eq:Js_final} is equivalent to minimizing the original acceleration-based objective \eqref{eq:naive_stiffness}, but yields a well-conditioned variational problem that is stable under discretization and suitable for gradient-based optimization.

This formulation differs fundamentally from existing approaches in that stiffness is optimized explicitly with respect to the dynamics actually integrated in stretched time, rather than inferred indirectly from solver step sizes or isolated geometric heuristics. As demonstrated in subsequent sections, the resulting time reparameterizations are smoother, less sensitive to parametric variation, and more compatible with learning-based reduced-order modeling frameworks such as neural ordinary differential equations.

\begin{algorithm}[h!]
\caption{Trajectory-Optimized Time Reparameterization}
\label{alg:proposed}
\begin{algorithmic}[1]
\Require Physical-time trajectory $\{(t_n,\bm{y}_n)\}_{n=0}^N$, target stretched final time $\tau_f$
\Ensure Reparameterized trajectory $\tilde{\bm{y}}(\tau)$ and time map $t(\tau)$ on a uniform $\tau$ grid

\State Compute discrete arc-length samples $\{s_n\}$ from $\bm{\varphi}$ and construct the arc-length trajectory $\bar{\bm{\varphi}}(s)$.
\State Interpolate $\bar{\bm{\varphi}}(s)$ onto a uniform arc-length grid $s\in[0,S]$.
\State Estimate the curvature $\kappa(s)$ of the trajectory along the arc-length coordinate.
\State Initialize an unconstrained speed function $v^*(s)$.
\State Enforce positivity of the speed via $v(s)=\exp(a+v^*(s))$, where $a$ is a scalar offset.
\State Determine $a$ such that the arrival-time constraint $\int_0^S v(s)^{-1}\,\mathrm{d}s=\tau_f$ is satisfied.
\State Solve the optimization problem
\[
\min_{v^*}
\int_0^S \left(v(s)v_s(s)^2+\kappa(s)^2\,v(s)^3\right)\,\mathrm{d}s,
\]
subject to the arrival-time constraint enforced through $a$.
\State Compute the stretched-time map $\tau(s)=\int_0^s v(\sigma)^{-1}\,\mathrm{d}\sigma$ and numerically invert it to obtain $s(\tau)$.
\State Construct the reparameterized trajectory $\tilde{\bm{y}}(\tau)=\bar{\bm{y}}(s(\tau))$ and the physical-time map $t(\tau)=t(s(\tau))$ by composing the two reparameterizations.
\State Interpolate $\tilde{\bm{y}}(\tau)$  and $t(\tau)$ onto uniform grid $\tau\in[0,\tau_f]$.
\end{algorithmic}
\end{algorithm}

Algorithm~\ref{alg:proposed} summarizes the proposed time reparameterization procedure. Given a physical-time trajectory, the method first constructs an arc-length representation to isolate geometric features of the solution independent of its original temporal parameterization. A smooth traversal speed along the arc-length coordinate is then obtained by solving a constrained optimization problem that balances tangential and curvature-induced acceleration while enforcing a prescribed final stretched time. The resulting speed profile defines a monotone time map, which is integrated and inverted to recover the reparameterized trajectory on a uniform stretched-time grid. By optimizing smoothness directly in the dynamics integrated in stretched time, the algorithm produces time reparameterizations that are stable under discretization and well suited for learning-based reduced-order modeling.

\section{Experimental Setup}
\label{sec:experimental}
\subsection{Training algorithm}

Network architectures are selected via a hyperparameter search based on Latin hypercube sampling (LHS). The sampled parameters include the network architecture, learning rate, the initial prediction horizon, and the interval at which the prediction horizon is increased during training. For each sampled configuration, a neural ODE model is trained. The LHS procedure is seeded to ensure that all three methods considered in this study are evaluated on an identical set of hyperparameter configurations. In total, each method is trained on the same set of $1000$ network pairs, with each pair consisting of a state network and a time-scaling network that are trained jointly.

The number of hidden layers for each network is sampled uniformly from $[1,10]$, while the number of learnable parameters per network is sampled in the range $[100,2500]$. The initial prediction horizon is varied between $2$ and $50$, and the horizon-increase frequency is sampled in the range $[1,10]$ epochs. Hyperparameter optimization is performed under a fixed computational budget of $1000$ network predictions per case per epoch. As a result, early training stages consist of many short predictions, whereas later epochs typically involve only one or two long-time integrations. Empirically, this approach led to improved training stability and enhanced model expressivity. The learning rate is sampled on a logarithmic scale by drawing $x \in [-5,-1]$ and setting $\mathrm{lr} = 5 \times 10^{x}$; it is subsequently decayed geometrically with epoch to reach $10^{-8}$ at the end of training. All models are pretrained for a fixed $1000$ iterations prior to a main training phase of $5000$ epochs.

The pretraining is done using forward Euler integration on single–time-step batches. This pretraining phase is computationally inexpensive and was found to promote stable optimization. Following pretraining, the networks are trained using the sampled learning rate and prediction horizon, with the horizon increased at intervals determined by the LHS. Forward Euler integration is used throughout the hyperparameter search to enable rapid evaluation of a large number of candidate configurations. Training and backpropagation through the time-integration scheme are performed using the \texttt{torchdiffeq} library~\cite{Chen2018}.

In implementation, both $f_{\mathrm{NN}}$ and $\alpha_{\mathrm{NN}}$ are represented using a composite network architecture. The right-hand side of the neural ODE is constructed by concatenating the outputs of two subnetworks, each of which receives as input the current system state with the parameter vector $\mu$ appended. All inputs are nondimensionalized to the interval $[-1,1]$ (based on the ranges encountered in the training data set) prior to evaluation. Similarly, the predicted derivatives are subsequently mapped back to physical units using scaling factors again derived from the minimum and maximum values of the derivatives observed in the on-reference data set. This normalization procedure improves training stability and ensures consistent scaling across and within systems.

Training is performed in stretched time $\tau$ using a mean-squared error objective,
\begin{equation}
\mathcal{L}_{\mathrm{MSE}}
=
\frac{1}{(N_q+1)N_\tau}
\left(
\sum_{i=1}^{N_q}\sum_{j=1}^{N_t}
\big(q_{i,j,\mathrm{pred}}-q_{i,j,\mathrm{true}}\big)^2
+
\sum_{j=1}^{N_\tau}
\big(t_{j,\mathrm{pred}}-t_{j,\mathrm{true}}\big)^2
\right),
\label{eq:MSE}
\end{equation} where $j$ indexes the step in $\tau$, and $N_q$ and $N_\tau$ are the number of state quantities and the number of steps in $\tau$ respectively. Equation~\eqref{eq:MSE} measures accuracy directly in the reparameterized coordinate. Prior to evaluation of~\eqref{eq:MSE}, all state variables are nondimensionalized such that the true data lie in the interval $[-1,1]$. The time coordinate is rescaled to the interval $[0,5]$, ensuring that comparable emphasis is placed on the state and time-scaling networks during training. This loss is used for training the neural ODE models.  As such, this loss is reparameterization dependent, and measures only the ability of a given network to produce the reference curves produced by the reparameterization.

Model selection and testing are performed using a reparameterization-invariant mean squared integral error,
\begin{equation}
\mathcal{L}_{\mathrm{MSIE}}
=
\frac{1}{N_q T}
\sum_{i=1}^{N_q}
\int_0^T
\left(
q_{i,\mathrm{pred}}(t)
-
q_{i,\mathrm{true}}(t)
\right)^2
\, \mathrm{d}t,
\label{eq:MSIE}
\end{equation}
where again $N_q$ is the number of state quantities and $T$ is the integration time. Equation~\eqref{eq:MSIE} is computed by trapezoidal integration in physical time. This metric evaluates all models ability to capture the same physical-time trajectories, independent of the underlying clock, enabling fair comparisons across reparameterization strategies.

\section{Results}
\label{sec:results}
\subsection{Time Reparameterization of Noisy Data}

Learning reduced-order models of stiff dynamical systems from high-fidelity simulations is fundamentally limited by the quality and structure of the training data. In many applications, full-order model data are generated by adaptive time integrators, and can be contaminated by numerical noise or sampled nonuniformly in ways that can depend on solver tolerances and parameter values. When such data are used directly to train neural ODE--based models, the learning process can become parameterized by these external effects rather than the underlying physics, potentially leading to ineffective distribution of $\tau$ resolution and difficult reparameterized ODEs. It therefor becomes important to generate reparameterized trajectories that are smooth in stretched time and stationary in parameter dependence. The ability to recover such smooth and stationary trajectories even from noisy FOM data helps construct robust, data-efficient neural ODE--based reduced-order models that faithfully represent the underlying dynamics.

Before assessing how the resulting trajectories impact neural ODE training, we first examine the ability of each reparameterization strategy to recover smooth and stationary clocks from noisy full-order data. To isolate this effect, we deliberately consider FOM trajectories generated using a high-order implicit Radau solver with very loose integration tolerances, resulting in time series that exhibit solver-directed noise and tolerance-dependent irregularity. This setting reflects a realistic scenario in which high-fidelity data might be contaminated by numerical artifacts. The goal of this portion of the study is therefore not to optimize accuracy with respect to a reference solution, but to evaluate how effectively each method extracts a coherent and tolerance-robust temporal structure from imperfect data.

\begin{figure}[h!]
\centering
\begin{subfigure}{0.32\textwidth}
  \centering
  \includegraphics[width=\linewidth]{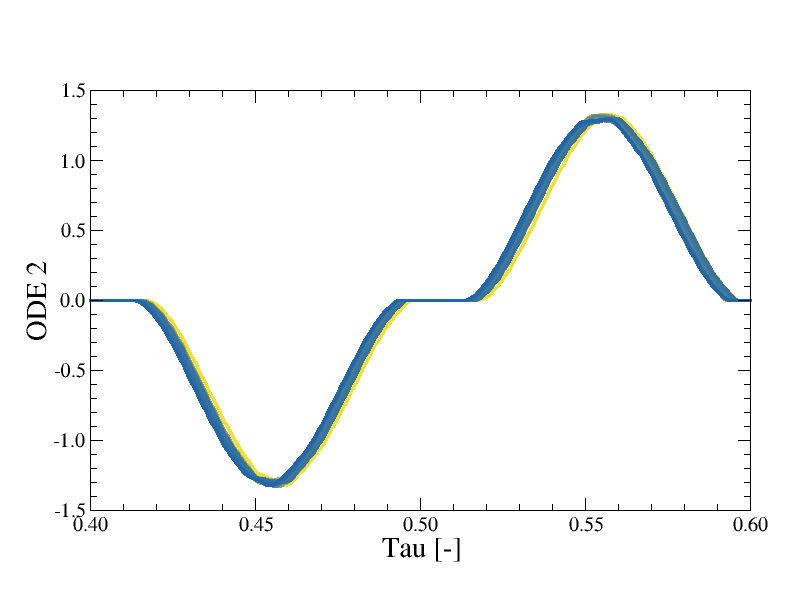}
  \caption{Proposed Method (TOTR)}
\end{subfigure}\hfill
\begin{subfigure}{0.32\textwidth}
  \centering
  \includegraphics[width=\linewidth]{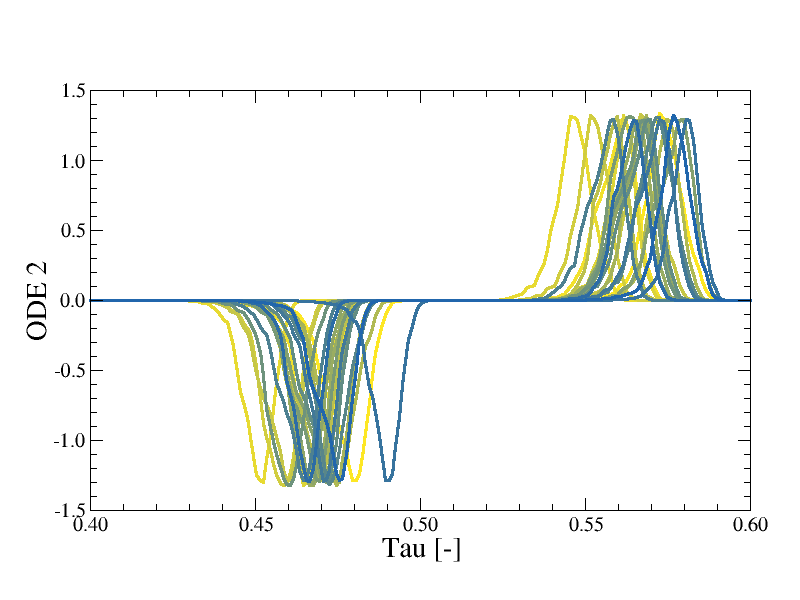}
  \caption{Solver Induced Method}
\end{subfigure}\hfill
\begin{subfigure}{0.32\textwidth}
  \centering
  \includegraphics[width=\linewidth]{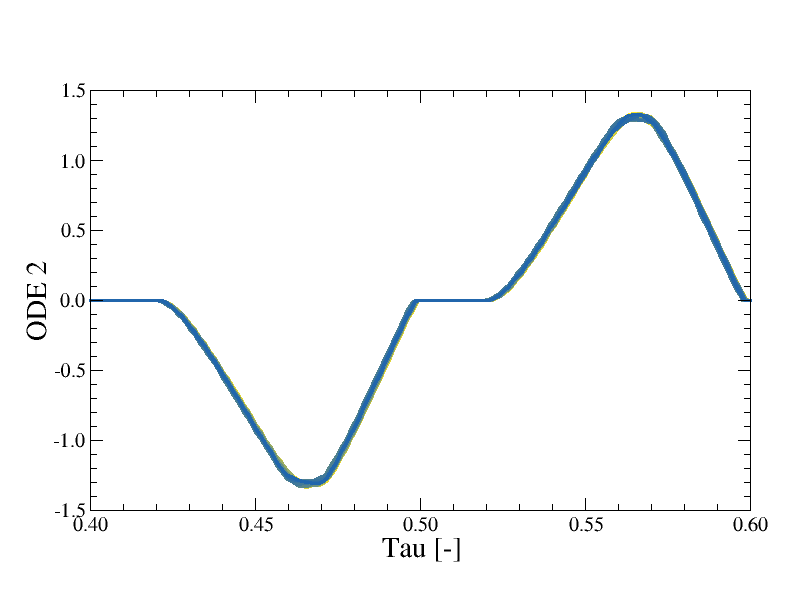}
  \caption{Extrema Based Method}
\end{subfigure}
\caption{Van der Pol oscillator $\mu\in[10^3,10^4]$ reparameterization by method for an integration tolerance of $10^{-3}$}
\label{fig:tol3}
\end{figure}

\begin{figure}[h!]
\centering
\begin{subfigure}{0.32\textwidth}
  \centering
  \includegraphics[width=\linewidth]{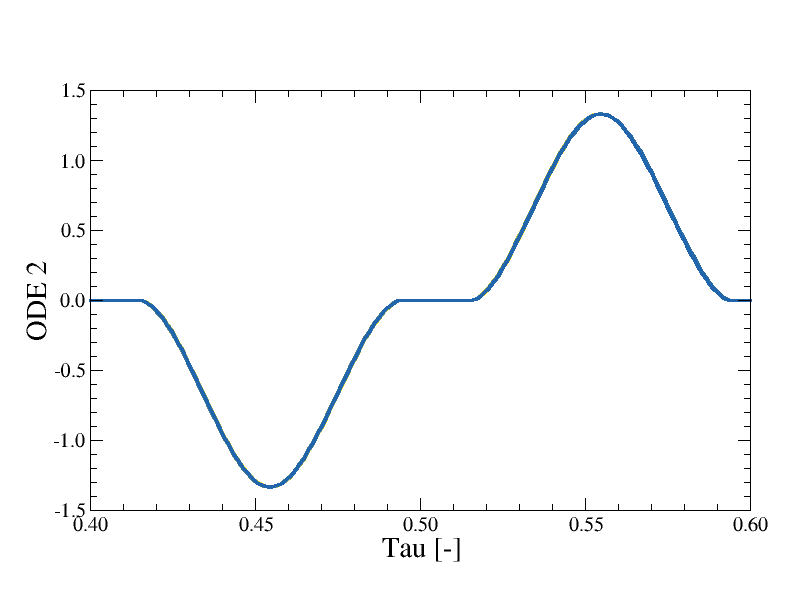}
  \caption{Proposed Method (TOTR)}
\end{subfigure}\hfill
\begin{subfigure}{0.32\textwidth}
  \centering
  \includegraphics[width=\linewidth]{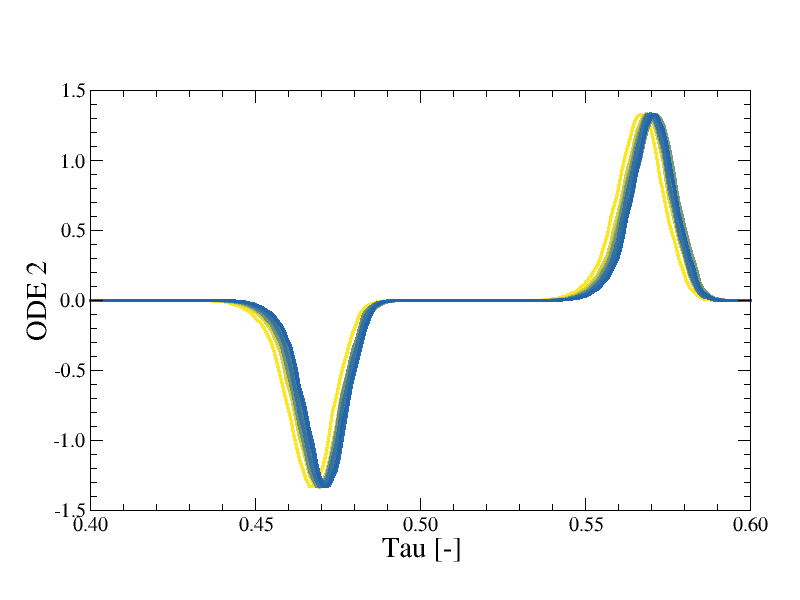}
  \caption{Solver Induced Method}
\end{subfigure}\hfill
\begin{subfigure}{0.32\textwidth}
  \centering
  \includegraphics[width=\linewidth]{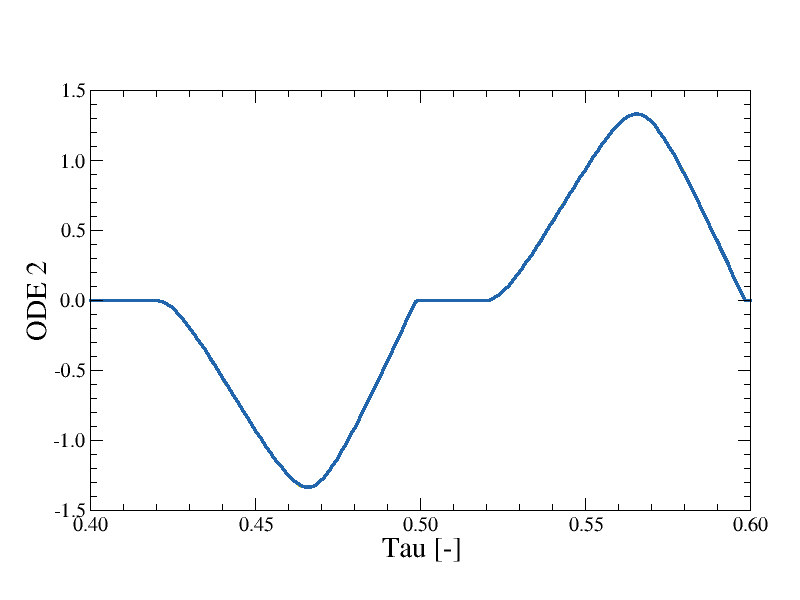}
  \caption{Extrema Based Method}
\end{subfigure}
\caption{Van der Pol oscillator $\mu\in[10^3,10^4]$ reparameterization by method for an integration tolerance of $10^{-6}$}
\label{fig:tol6}
\end{figure}

Figures~\ref{fig:tol3} and~\ref{fig:tol6} illustrate these issues for the van der Pol oscillator later considered in Section~\ref{sec:results_vdp} over $\mu\in[10^3,10^4]$, comparing three time reparameterization strategies at integration tolerances of $10^{-3}$ and $10^{-6}$. The solver-directed reparameterization naturally exhibits pronounced sensitivity to the integration tolerance, with substantial changes in the resulting time map as the tolerance is reduced. This behavior arises because the clock is inherited from adaptive solver step-size control and therefore reflects numerical parameters rather than intrinsic properties of the dynamics. As a result, the same set of physical trajectories can yield markedly different stretched-time representations depending solely on solver settings, undermining the consistency required for effective learning.

In contrast, both the proposed trajectory-optimized method and the extrema-based method produce substantially more stationary and repeatable reparameterized trajectories across tolerances, yielding far more uniform stretched-time representations. These methods construct clocks that are tied directly to the structure of the trajectory itself rather than to numerical integration details, resulting in time maps that are significantly more robust to solver noise and tolerance variation.

It is important to emphasize that the Van der Pol oscillator considered here lies in an asymptotic relaxation regime for large $\mu$, where the separation between slow and fast dynamics is well established. In this limit, it is possible to extract a stretched-time parametrization that is effectively stationary with respect to $\mu$, as evidenced by the consistent time maps observed across the parameter range shown. This favorable structure should not be interpreted as a generic property of stiff dynamical systems. In many systems of practical interest, the stiffness structure evolves with parameters, and reparameterizations that are both smooth and parameter-independent may not exist. Accordingly, the highly uniform time maps in Figures~\ref{fig:tol3} and~\ref{fig:tol6} should be viewed as an example of achievable regularity rather than a universal outcome.

Nevertheless, a key observation is that although the proposed method explicitly optimizes smoothness in stretched time $\tau$, the resulting reparameterizations also exhibit a high degree of smoothness with respect to the parameter $\mu$. This emergent joint smoothness in $\tau$ and $\mu$ is particularly valuable for neural ODE training, as it promotes coherent trajectories across parameter space, reduces parameter-induced variability, and improves the stability and generalization of learned reduced-order models.
\subsection{Problem 1: Parameterized Stiff Linear System}

The first problem considered is a parameterized linear system constructed to isolate the effects of stiffness while avoiding nonlinear geometric complexity. The governing equations are
\begin{equation}
\deriv{\bm{y}}{t}
=
\bm{A}\,\bm{B}(\mu)\,\bm{A}^{\top}\bm{y}(t;\mu),
\qquad
\bm{y}(0)=
\begin{bmatrix}
1 & 0 & 0 & 0 & 0
\end{bmatrix}^{\top},
\label{eq:sls}
\end{equation}
where \(\bm{y}(t;\mu)\in\mathbb{R}^5\) and \(\mu>0\) controls stiffness.

The matrix \(\bm{A}\in\mathbb{R}^{5\times 5}\) is an orthogonal mixing matrix obtained via a QR factorization of a fixed dense matrix,
\begin{equation}
\bm{A}
=
\mathrm{qr}\!\left(
\begin{bmatrix}
 5 &  1 & -1 &  0 &  0\\
-1 &  3 &-10 &  1 &  2\\
 2 & -1 &  5 &  1 & -1\\
 0 &  1 &  2 &  3 & -3\\
 1 & 12 & -1 & -2 &  5
\end{bmatrix}
\right),
\label{eq:sls_A}
\end{equation}
which deliberately mixes fast and slow modes across all state components. This ensures that stiffness is not confined to a single coordinate direction but instead appears in each component of \(\bm{y}\), providing a stringent test for time reparameterization.

The parameter-dependent decay rates are contained in
\begin{equation}
\bm{B}(\mu)
=
\begin{bmatrix}
-0.25\mu & 0.97\mu & 0 & 0 & 0\\
-0.97\mu &-0.25\mu & 0 & 0 & 0\\
0 & 0 & -\dfrac{10\mu}{\mu^{0.3}+10} & 0 & 0\\
0 & 0 & 0 & -\dfrac{100\mu}{\mu^{0.6}+100} & 0\\
0 & 0 & 0 & 0 & -\dfrac{\mu}{\mu+1}
\end{bmatrix}.
\label{eq:sls_B}
\end{equation}
The dominant decay rates scale as \(\mathcal{O}(\mu)\), while the slowest mode remains \(\mathcal{O}(1)\) as $\mu$ increases. Consequently, the stiffness ratio grows proportionally to \(\mu\).

Training data were generated for
\[
\mu \in \{10^{1},10^{1.1},10^{1.2},\ldots,10^{3.9},10^{4}\},
\]
with off-reference testing performed at intermediate parameter values
\[
\mu_{\mathrm{test}} \in \{10^{1.05},10^{1.65},10^{2.25},10^{2.75},10^{3.35},10^{3.95}\}.
\]
This stiff linear system (SLS) provides a controlled setting in which the effect of time reparameterization can be studied independently of nonlinear dynamics. Because the system contains five state variables, only a representative subset is shown for clarity. The results for ODEs $y_1$ and $y_4$ are presented in the following plots. The variable $y_1$ was selected because it represents the perturbed quantity, while $y_4$ was chosen because it exhibits the largest error across the methods considered.

\begin{figure}[h!]
\centering
\begin{subfigure}{0.5\textwidth}
  \centering
  \includegraphics[width=\linewidth]{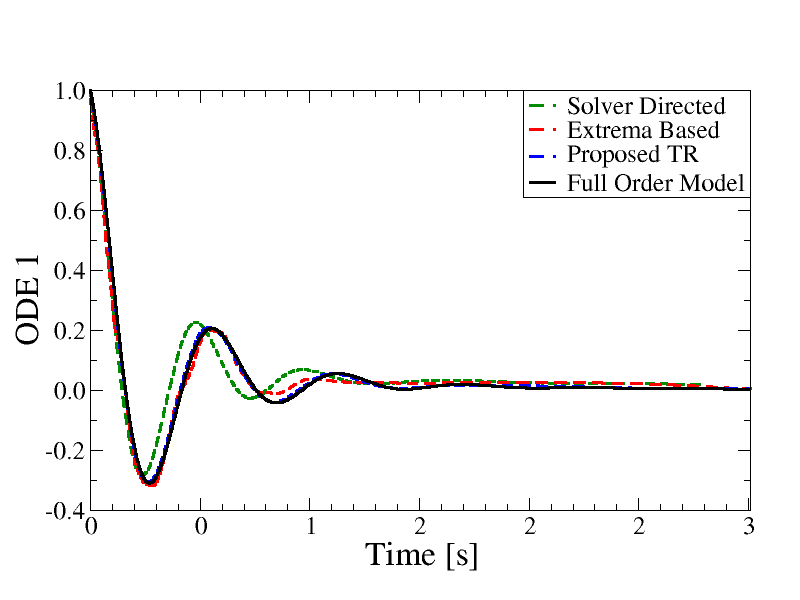}
  \caption{ODE one}
\end{subfigure}\hfill
\begin{subfigure}{0.5\textwidth}
  \centering
  \includegraphics[width=\linewidth]{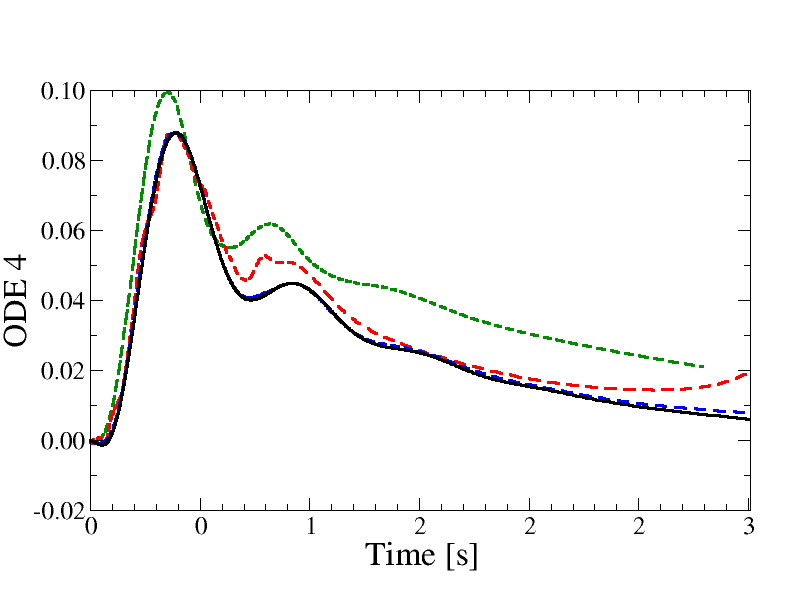}
  \caption{ODE four}
\end{subfigure}\hfill
\caption{Time reparameterizations for the stiff linear system, off-reference test case
\(\mu=10^{1.05}\).}
\label{fig:sls_case2a}
\end{figure}

\paragraph{Moderate stiffness regime.}
Figure~\ref{fig:sls_case2a} compares reparameterizations for the off-reference test case \(\mu=10^{1.05}\). At this value of $\mu$ very little stiffness is present. All three methods capture the trends present with varying degrees of success. The proposed method captures the dynamics with very little error. The extrema-based method captures the general trend of the solution, but introduces spurious oscillations due to the trajectory in the reparameterized coordinate, $\tau$. The solver-directed does poorly here, as the behavior of the problem in the stiff regime inhibits the ability of the networks to learn effectively in this regime.

\begin{figure}[h!]
\centering
\begin{subfigure}{0.5\textwidth}
  \centering
  \includegraphics[width=\linewidth]{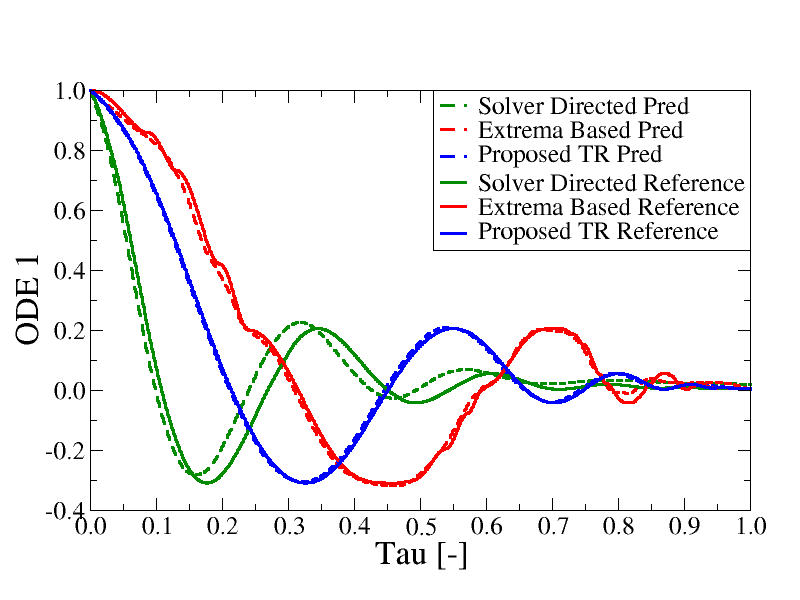}
  \caption{ODE one}
\end{subfigure}\hfill
\begin{subfigure}{0.5\textwidth}
  \centering
  \includegraphics[width=\linewidth]{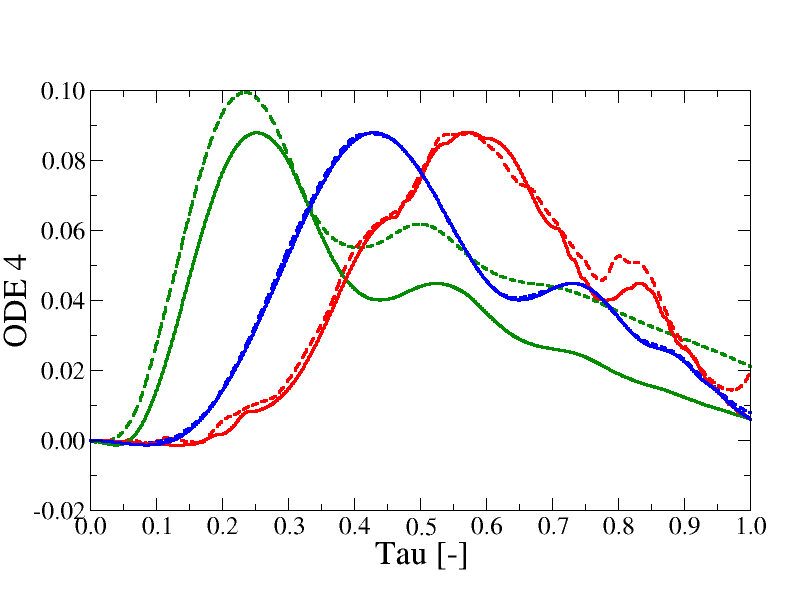}
  \caption{ODE four}
\end{subfigure}\hfill
\caption{Trajectories for \(\mu=10^{1.05}\) in the reparameterized coordinate, $\tau$. }
\label{fig:sls_case2b}
\end{figure}

Figure~\ref{fig:sls_case2b} compares the trajectories of the solutions in $\tau$. Though the solution does not exhibit much stiffness, all three methods slow down time in areas of larger gradients.  The proposed method  is able to match the reference trajectory more accurately than the benchmarks. The extrema-based method is able to predict the the reference solution with reasonable accuracy, but the method's trajectory in $\tau$ is more difficult for the network to approximate. 

\begin{figure}[h!]
\centering
\begin{subfigure}{0.5\textwidth}
  \centering
  \includegraphics[width=\linewidth]{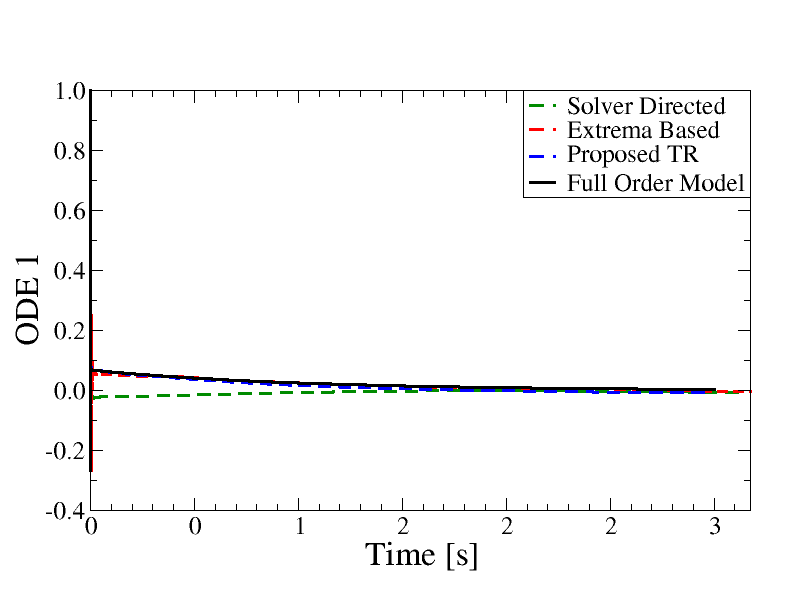}
  \caption{ODE one}
\end{subfigure}\hfill
\begin{subfigure}{0.5\textwidth}
  \centering
  \includegraphics[width=\linewidth]{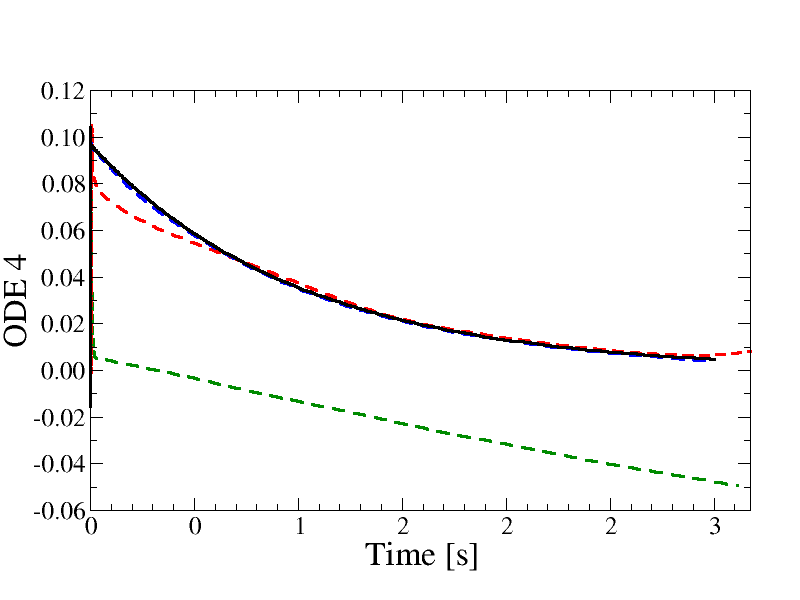}
  \caption{ODE four}
\end{subfigure}\hfill
\caption{Time reparameterizations for the stiff linear system, off-reference test case
\(\mu=10^{3.95}\).}
\label{fig:sls_case36a}
\end{figure}

\begin{figure}[h!]
\centering
\begin{subfigure}{0.5\textwidth}
  \centering
  \includegraphics[width=\linewidth]{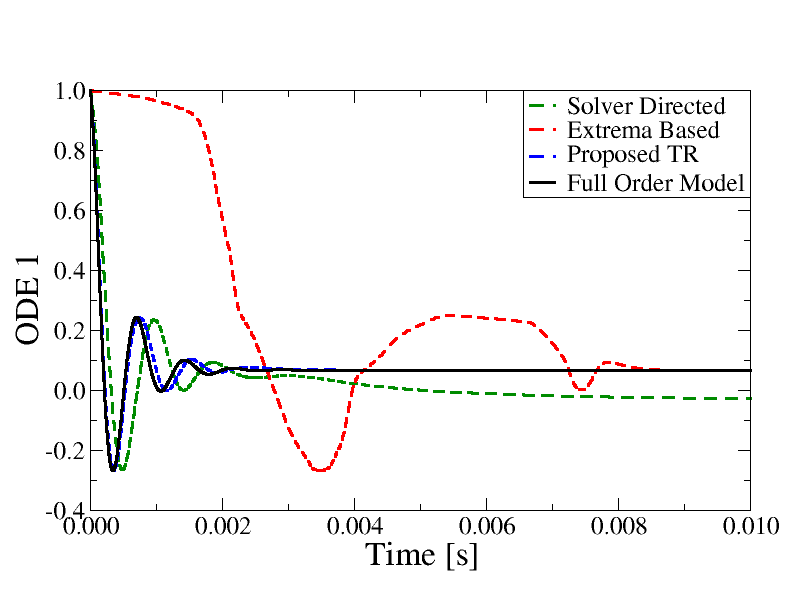}
  \caption{ODE one}
\end{subfigure}\hfill
\begin{subfigure}{0.5\textwidth}
  \centering
  \includegraphics[width=\linewidth]{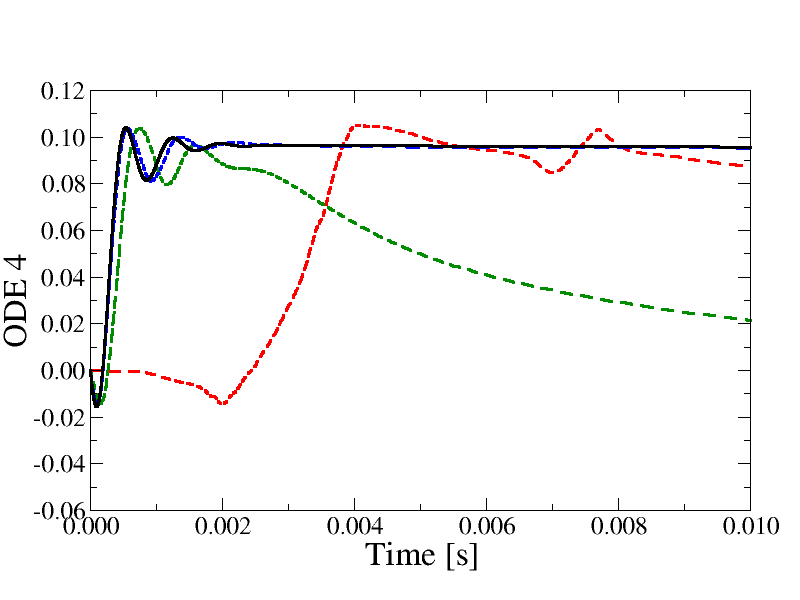}
  \caption{ODE four}
\end{subfigure}\hfill
\caption{Early-time zoom (first ten milliseconds) for \(\mu=10^{3.95}\), highlighting
resolution of the fastest transient scales.}
\label{fig:sls_case36b}
\end{figure}

\paragraph{Extreme stiffness regime.}
As stiffness increases, differences between reparameterization strategies become more pronounced. Figures~\ref{fig:sls_case36a} and~\ref{fig:sls_case36b} show two representative views for \(\mu=10^{3.95}\). The proposed method maintains a smooth time map that resolves both fast and slow dynamics. In contrast, the solver-directed method under-resolves slow-time recovery, while the  extrema-based heuristic fails to adequately resolve the fastest transient scales, as highlighted in the early-time zoom of Figure~\ref{fig:sls_case36b}.

\begin{figure}[h!]
\centering
\begin{subfigure}{0.5\textwidth}
  \centering
  \includegraphics[width=\linewidth]{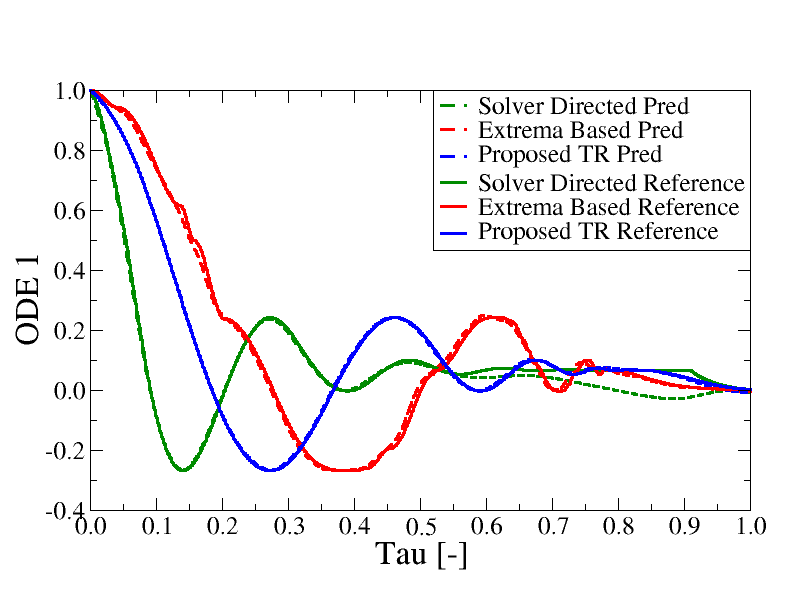}
  \caption{ODE one}
\end{subfigure}\hfill
\begin{subfigure}{0.5\textwidth}
  \centering
  \includegraphics[width=\linewidth]{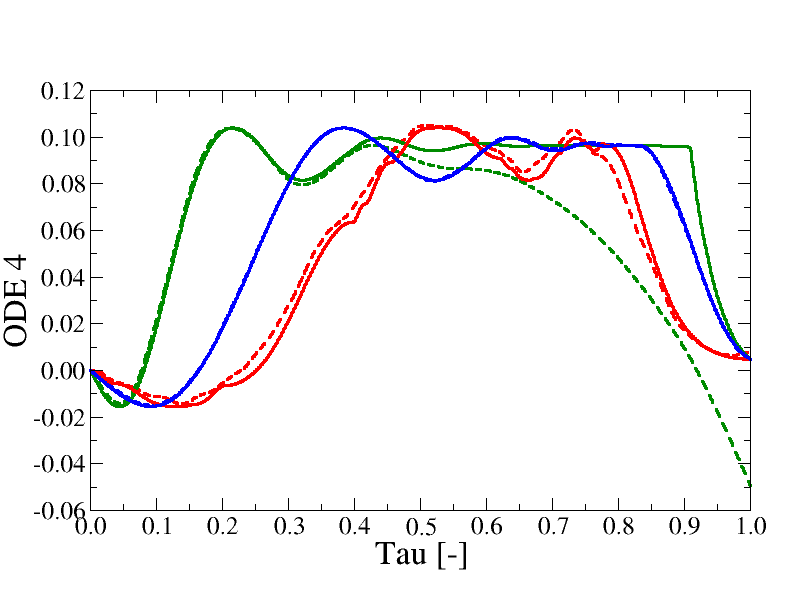}
  \caption{ODE four}
\end{subfigure}\hfill
\caption{Trajectories for \(\mu=10^{3.95}\) in the reparameterized coordinate, $\tau$ }
\label{fig:sls_case36c}
\end{figure}

Figure~\ref{fig:sls_case36c} elucidates the mechanism behind the failure of the solver-directed method. The trajectories produced in $\tau$ have introduced a cusp-like behavior at the transition between the two time scales. The neural networks failed to adequately capture this behavior. 

\begin{figure}[h!]
\centering
\begin{subfigure}{0.5\textwidth}
  \centering
  \includegraphics[width=\linewidth]{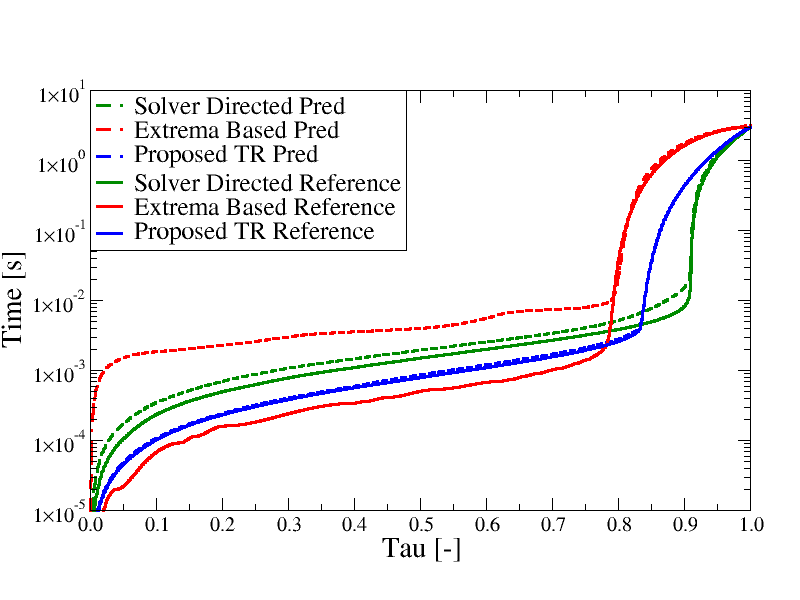}
  \caption{Comparison of time as a function of $\tau$ between the three methods}
\end{subfigure}\hfill
\begin{subfigure}{0.5\textwidth}
  \centering
  \includegraphics[width=\linewidth]{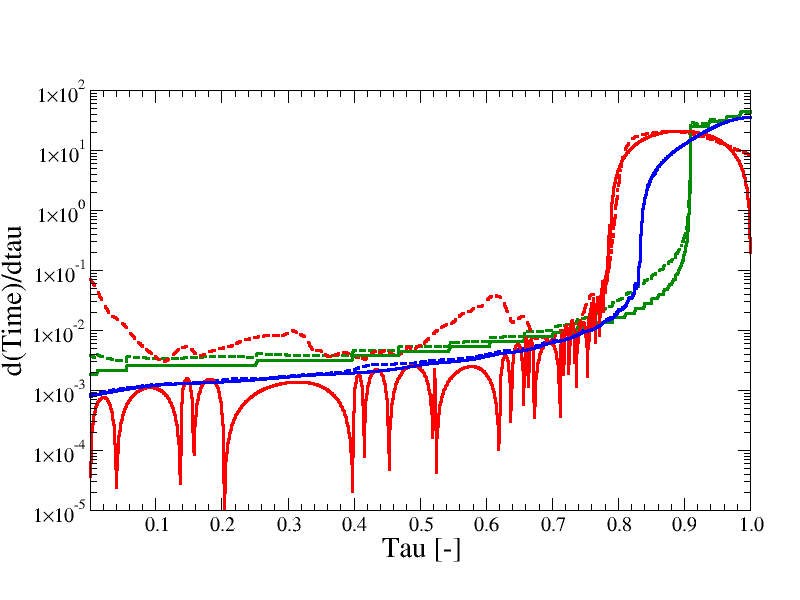}
  \caption{Comparison of the derivative of each methods time with respect to $\tau$.}
\end{subfigure}\hfill
\caption{Time curves and their derivatives for \(\mu=10^{3.95}\). }
\label{fig:sls_case36d}
\end{figure}

The difference between the methods becomes clearer when the different reparameterizations are compared directly. In Figure~\ref{fig:sls_case36d} the time curves are compared, as well as their derivatives. Even though the solver-directed method introduced stiff behavior in the reparameterized ODEs, the time curve itself is able to be represented with fair accuracy.  In contrast, while the extrema-based method produced ODEs that were more easily captured with a neural network, the time curve required to do so was extremely difficult for the network to match. In particular, the neural network $ \alpha_{\mathrm{NN}}(\tilde{\bm{z}},\tau;\mu)$ must predict at each step the derivative of time with respect to $\tau$.  The time curve constructed for this case varies wildly, and the network fails to capture the quick oscillations present in the first three-quarters of the trajectory. It can be observed that for this case the proposed method results in smoother time curves that are more easily represented by ML-ROMs. Furthermore, the resulting reparameterized ODEs are not only smoother in $\tau$, but by comparing Figures~\ref{fig:sls_case2b} and~\ref{fig:sls_case36c} are more stationary with respect to variations in $\mu$. These together result in a modeling problem that is a much simpler task for the NN-ROM to capture.

\paragraph{Discussion of problem one results.}
Because the underlying ODE, integration tolerances, and training procedure are held fixed, all differences observed in Figures~\ref{fig:sls_case2a}--\ref{fig:sls_case36d} arise solely from the choice of time reparameterization. The results indicate that directly optimizing the clock to promote smooth trajectories in stretched time yields more regular time maps, particularly in high-stiffness regimes. This regularity is advantageous for parametric ROM construction: small variations in the state should not induce disproportionately large changes in the inferred clock, and smoother maps are typically easier to learn robustly within a data-driven surrogate.

\begin{table}[htbp]
\small
\centering
\caption{SLS $\tau$-MSE for all methods (On- vs Off-reference).}
\label{tab:sls_tau_mse_nondim}
\setlength{\tabcolsep}{6pt}
\renewcommand{\arraystretch}{1.15}
\begin{tabular}{llcccccc}
\toprule
& & \multicolumn{2}{c}{\textbf{Solver-Directed}} & \multicolumn{2}{c}{\textbf{Extrema-Based}} & \multicolumn{2}{c}{\textbf{Trajectory-Optimized}} \\
\cmidrule(lr){3-4}\cmidrule(lr){5-6}\cmidrule(lr){7-8}
\textbf{Block} & \textbf{Metric} & \textbf{ON} & \textbf{OFF} & \textbf{ON} & \textbf{OFF} & \textbf{ON} & \textbf{OFF} \\
\midrule
\multirow{8}{*}{\makecell[l]{$\tau$ MSE\\(nondim)}} 
& $y_1$  & $5.99\times10^{-4}$ & $8.65\times10^{-4}$ & $2.33\times10^{-4}$ & $2.54\times10^{-4}$ & $3.74\times10^{-5}$ & $4.82\times10^{-5}$ \\
& $y_2$  & $8.51\times10^{-3}$ & $1.23\times10^{-2}$ & $5.95\times10^{-4}$ & $6.92\times10^{-4}$ & $4.00\times10^{-5}$ & $5.72\times10^{-5}$ \\
& $y_3$  & $4.80\times10^{-3}$ & $6.62\times10^{-3}$ & $2.73\times10^{-4}$ & $3.15\times10^{-4}$ & $1.40\times10^{-4}$ & $1.98\times10^{-4}$ \\
& $y_4$  & $1.50\times10^{-2}$ & $2.07\times10^{-2}$ & $5.63\times10^{-4}$ & $7.32\times10^{-4}$ & $4.44\times10^{-4}$ & $6.42\times10^{-4}$ \\
& $y_5$  & $8.87\times10^{-4}$ & $1.37\times10^{-3}$ & $3.67\times10^{-4}$ & $4.12\times10^{-4}$ & $4.38\times10^{-5}$ & $6.03\times10^{-5}$ \\
& Time   & $2.25\times10^{-4}$ & $2.83\times10^{-4}$ & $1.62\times10^{-4}$ & $1.68\times10^{-4}$ & $4.41\times10^{-6}$ & $8.23\times10^{-6}$ \\
& State  & $5.96\times10^{-3}$ & $8.36\times10^{-3}$ & $4.06\times10^{-4}$ & $4.81\times10^{-4}$ & $1.41\times10^{-4}$ & $2.01\times10^{-4}$ \\
& All    & $5.00\times10^{-3}$ & $7.02\times10^{-3}$ & $3.65\times10^{-4}$ & $4.29\times10^{-4}$ & $1.18\times10^{-4}$ & $1.69\times10^{-4}$ \\
\bottomrule
\end{tabular}
\end{table}

\begin{table}[htbp]
\small
\centering
\caption{SLS MSIE (dimensional) for all methods (On- vs Off-reference).}
\label{tab:sls_msie_dim}
\setlength{\tabcolsep}{6pt}
\renewcommand{\arraystretch}{1.15}
\begin{tabular}{llcccccc}
\toprule
& & \multicolumn{2}{c}{\textbf{Solver-Directed}} & \multicolumn{2}{c}{\textbf{Extrema-Based}} & \multicolumn{2}{c}{\textbf{Trajectory-Optimized}} \\
\cmidrule(lr){3-4}\cmidrule(lr){5-6}\cmidrule(lr){7-8}
\textbf{Block} & \textbf{Metric} & \textbf{ON} & \textbf{OFF} & \textbf{ON} & \textbf{OFF} & \textbf{ON} & \textbf{OFF} \\
\midrule
\multirow{6}{*}{\makecell[l]{MSIE\\(dim)}} 
& $y_1$  & $6.70\times10^{-4}$ & $9.46\times10^{-4}$ & $2.65\times10^{-4}$ & $2.96\times10^{-4}$ & $1.07\times10^{-4}$ & $1.42\times10^{-4}$ \\
& $y_2$  & $8.49\times10^{-4}$ & $1.32\times10^{-3}$ & $3.68\times10^{-5}$ & $3.75\times10^{-5}$ & $5.97\times10^{-6}$ & $8.88\times10^{-6}$ \\
& $y_3$  & $1.65\times10^{-3}$ & $2.61\times10^{-3}$ & $7.64\times10^{-5}$ & $8.12\times10^{-5}$ & $8.94\times10^{-5}$ & $1.35\times10^{-4}$ \\
& $y_4$  & $4.00\times10^{-4}$ & $6.23\times10^{-4}$ & $1.54\times10^{-5}$ & $1.78\times10^{-5}$ & $2.47\times10^{-5}$ & $3.83\times10^{-5}$ \\
& $y_5$  & $3.38\times10^{-4}$ & $6.31\times10^{-4}$ & $1.87\times10^{-4}$ & $1.85\times10^{-4}$ & $1.11\times10^{-5}$ & $1.43\times10^{-5}$ \\
& State  & $7.81\times10^{-4}$ & $1.23\times10^{-3}$ & $1.16\times10^{-4}$ & $1.23\times10^{-4}$ & $4.77\times10^{-5}$ & $6.77\times10^{-5}$ \\
\bottomrule
\end{tabular}
\end{table}

Tables~\ref{tab:sls_tau_mse_nondim} and~\ref{tab:sls_msie_dim} demonstrate that the proposed time-reparameterization strategy consistently outperforms the comparison methods in both stretched-time accuracy and physical-time trajectory prediction. The proposed method achieves reductions of one to two orders of magnitude in $\tau$-MSE across all state components and in the aggregated time and state metrics, for both on- and off-reference evaluations. These improvements indicate that directly optimizing the clock to promote smooth dynamics in stretched time yields substantially more regular and robust time maps that are more easily represented by ML-ROMs.  Notably, we see that the performance advantage persists under off-reference testing, suggesting that all the learned reparameterizations generalize beyond the training trajectory and are not overfit to a specific reference configuration. Taken together, these results demonstrate that the smoother trajectories of the proposed approach yield time maps that are easier to learn.

\subsection{Problem 2: Van der Pol Oscillator}
\label{sec:results_vdp}
The second benchmark problem is the Van der Pol oscillator, a classical nonlinear system that exhibits strong stiffness in the relaxation regime. Written as a first-order system,
\begin{subequations}
\begin{align}
\dot{y}_1 &= y_2, \\
\dot{y}_2 &= \mu^2 \left((1 - y_1^2) y_2 - y_1\right),
\end{align}
\end{subequations}
where \(\mu>0\) controls the stiffness. For \(\mu\gg 1\), solutions alternate between slow evolution along a manifold and rapid transitions across thin boundary layers, yielding an effective stiffness ratio that scales as \(\mathcal{O}(\mu)\).

Training data were generated for
\[
\mu \in \{10^{2},10^{2.04},10^{2.08},\ldots,10^{3.96},10^{4}\},
\]
with off-reference testing performed at
\[
\mu_{\mathrm{test}} \in \{10^{2.01},10^{2.67},10^{3.33},10^{3.99}\}.
\]

\begin{figure}[h!]
\centering
\begin{subfigure}{0.5\textwidth}
  \centering
  \includegraphics[width=\linewidth]{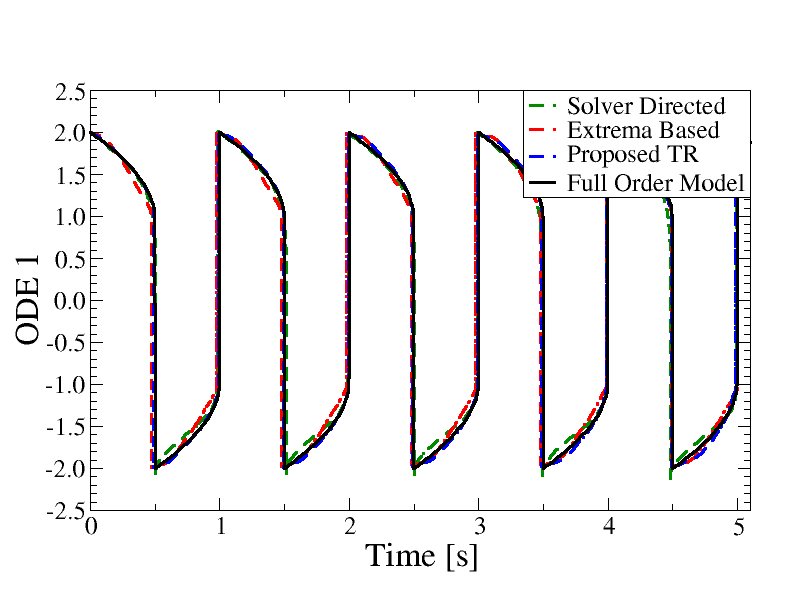}
  \caption{ODE one}
\end{subfigure}\hfill
\begin{subfigure}{0.5\textwidth}
  \centering
  \includegraphics[width=\linewidth]{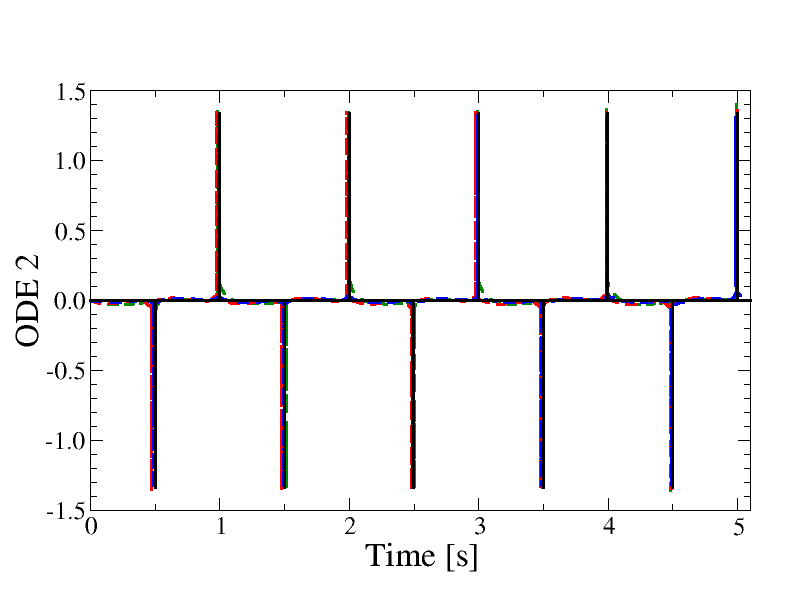}
  \caption{ODE two}
\end{subfigure}\hfill
\caption{Van der Pol oscillator, off-reference test case \(\mu=10^{2.01}\).}
\label{fig:vdp_case2_a}
\end{figure}

\paragraph{Moderate stiffness regime.}
Figure~\ref{fig:vdp_case2_a} compares time reparameterizations for \(\mu=10^{2.01}\). All three approaches spread fast transitions over a wider interval in \(\tau\), enabling explicit integration on a uniform stretched-time grid. In this regime, while differences are primarily qualitative, the proposed method not only exhibits improved adherence to the slow manifold over the full cycle, but best captures the behaviour of the boundary layer. 

\begin{figure}[h!]
\centering
\begin{subfigure}{0.5\textwidth}
  \centering
  \includegraphics[width=\linewidth]{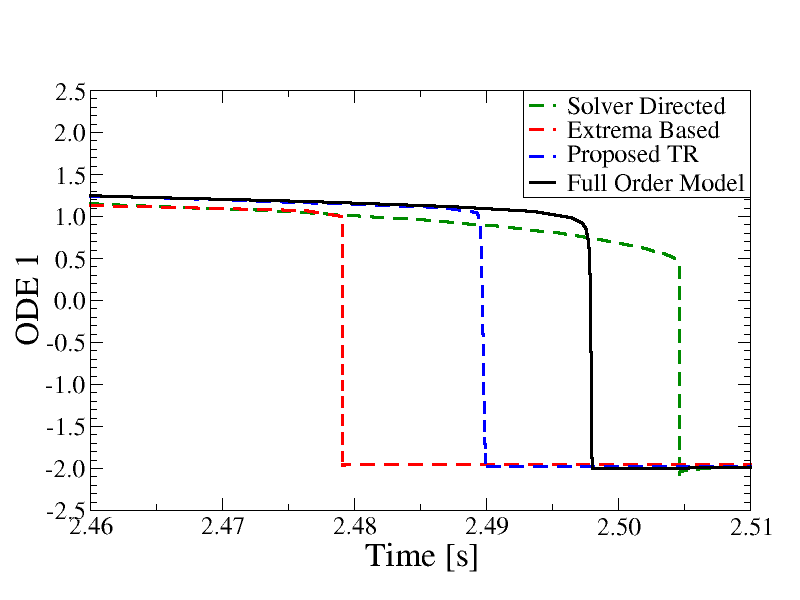}
  \caption{ODE one}
\end{subfigure}\hfill
\begin{subfigure}{0.5\textwidth}
  \centering
  \includegraphics[width=\linewidth]{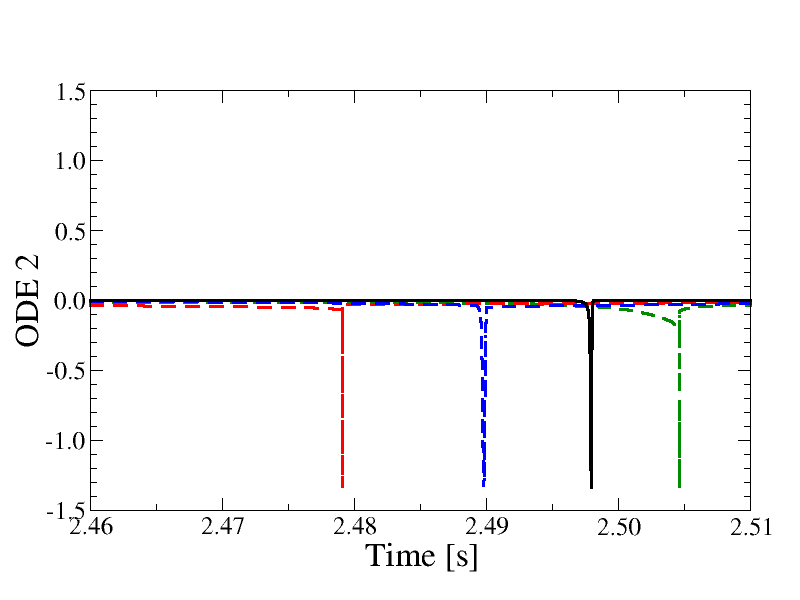}
  \caption{ODE two}
\end{subfigure}\hfill
\caption{Van der Pol oscillator, off-reference test case \(\mu=10^{2.01}\), emphasizing behavior near a boundary layer.}
\label{fig:vdp_case2_b}
\end{figure}

Figure~\ref{fig:vdp_case2_b} provides an alternate visualization of the same case, highlighting differences in the performance near the boundary layer. All three methods have phase error of the same magnitude. 

\begin{figure}[h!]
\centering
\begin{subfigure}{0.33\textwidth}
  \centering
  \includegraphics[width=\linewidth]{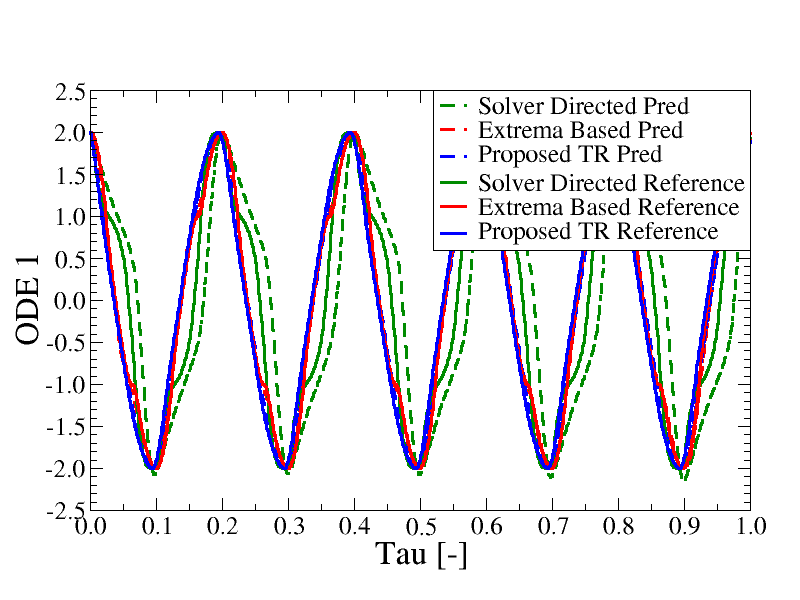}
  \caption{ODE one}
\end{subfigure}\hfill
\begin{subfigure}{0.33\textwidth}
  \centering
  \includegraphics[width=\linewidth]{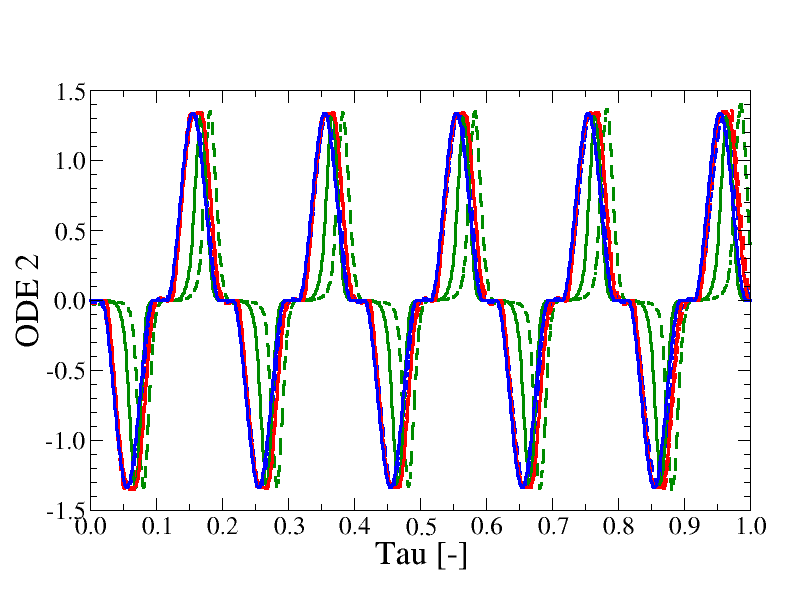}
  \caption{ODE two}
\end{subfigure}\hfill
\begin{subfigure}{0.33\textwidth}
  \centering
  \includegraphics[width=\linewidth]{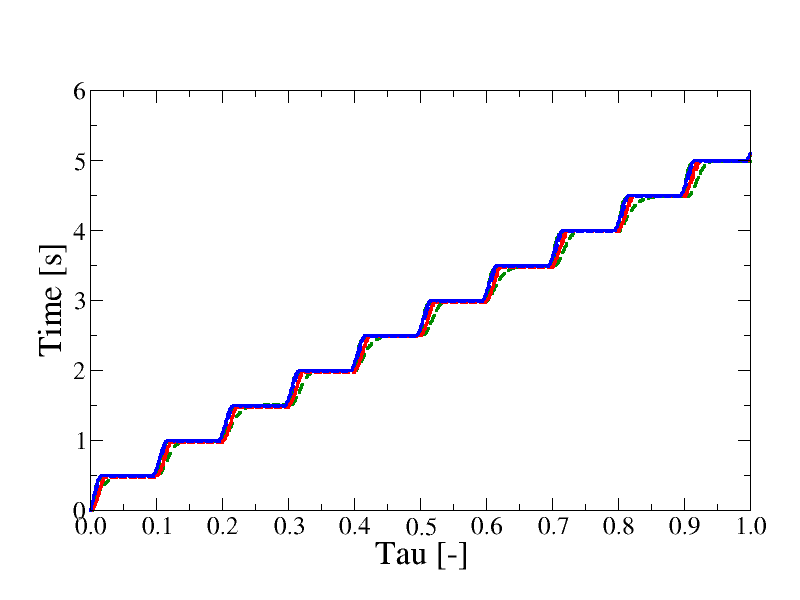}
  \caption{Time vs. $\tau$}
\end{subfigure}\hfill
\caption{Van der Pol oscillator, off-reference test case \(\mu=10^{2.01}\), trajectories in the reparameterized coordinate.}
\label{fig:vdp_case2_c}
\end{figure}

Figure~\ref{fig:vdp_case2_c} provides the trajectories in $\tau$. The proposed method and the extrema-based method produced similar trajectories. This is anticipated as both methods share the common arc-length trial parametrization, and both methods will stretch time in similar places as the extrema are located near the stiff events of the system. The solver-directed method experiences phase lag in both the state and time predictions. Despite this error, the method produced a solution in time that is still remarkably accurate and, of the three methods tested, produced the least phase error in time.

\begin{figure}[h!]
\centering
\begin{subfigure}{0.5\textwidth}
  \centering
  \includegraphics[width=\linewidth]{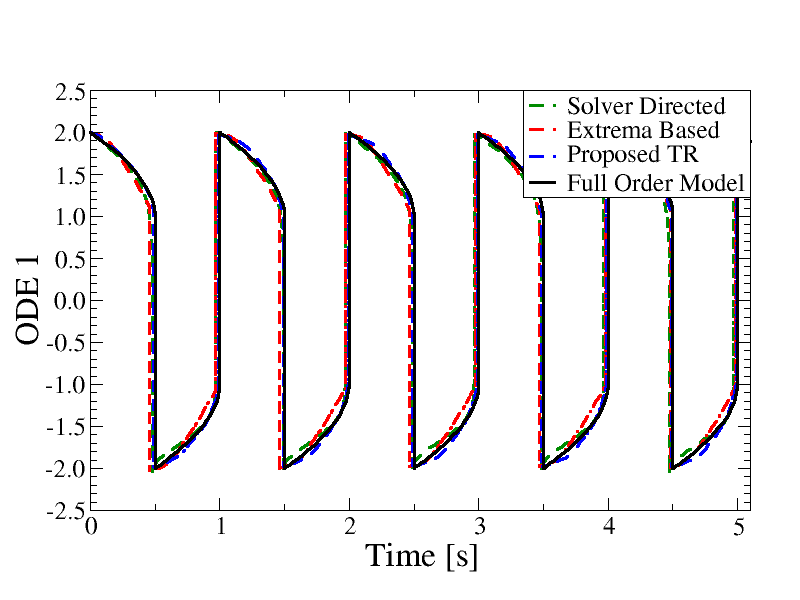}
  \caption{ODE one }
\end{subfigure}\hfill
\begin{subfigure}{0.5\textwidth}
  \centering
  \includegraphics[width=\linewidth]{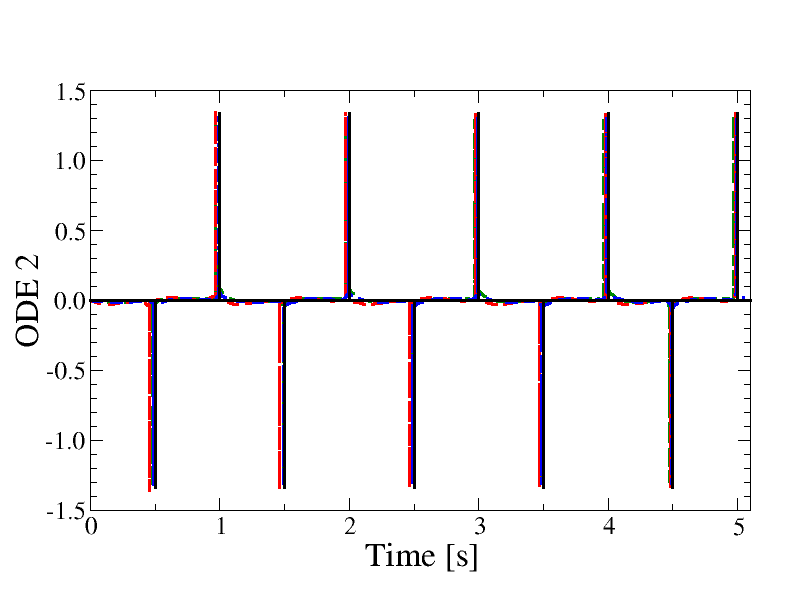}
  \caption{ODE two}
\end{subfigure}\hfill
\caption{Van der Pol oscillator, off-reference test case \(\mu=10^{3.99}\):
comparison of time reparameterizations in the extreme stiffness regime.}
\label{fig:vdp_case54a}
\end{figure}

\paragraph{Extreme stiffness regime.}
Figure~\ref{fig:vdp_case54a} shows results for \(\mu=10^{3.99}\). In this regime the three
methods exhibit clearer tradeoffs: all reduce stiffness enough to permit explicit
integration, but they differ in how they balance phase accuracy, manifold tracking, and
resolution of rapid transitions. In these particular results, the solver-directed
strategy tracks the slow manifold most accurately, while the other approaches more
faithfully capture the shape of the fast transitions at the cost of increased phase lead.

\begin{figure}[h!]
\centering
\begin{subfigure}{0.5\textwidth}
  \centering
  \includegraphics[width=\linewidth]{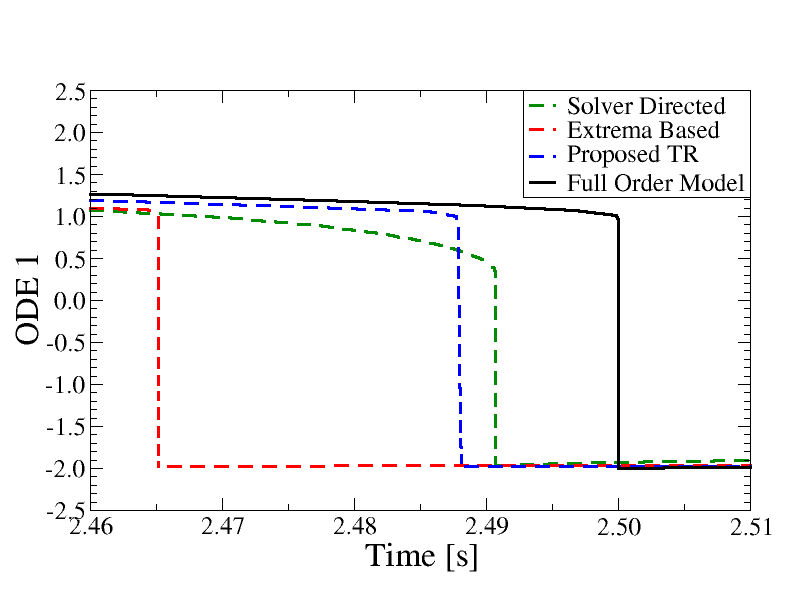}
  \caption{ODE one}
\end{subfigure}\hfill
\begin{subfigure}{0.5\textwidth}
  \centering
  \includegraphics[width=\linewidth]{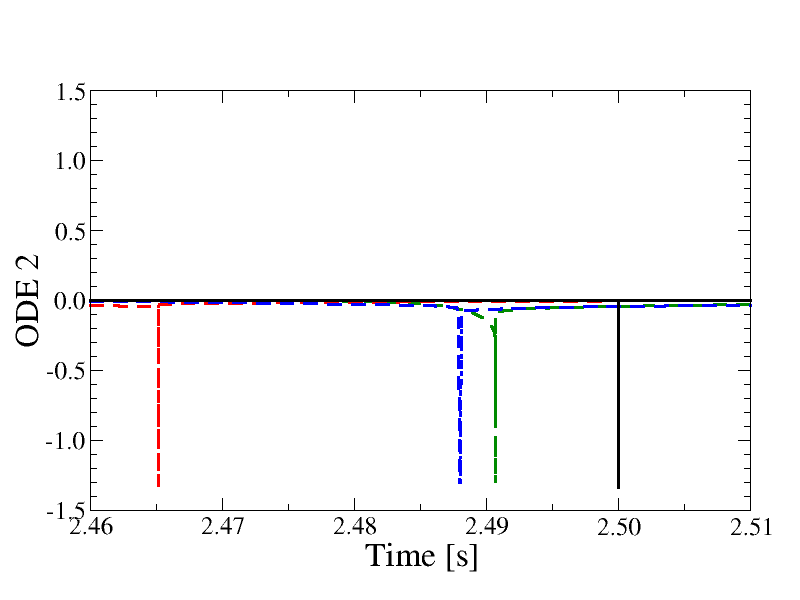}
  \caption{ODE two}
\end{subfigure}\hfill
\caption{Van der Pol oscillator, off-reference test case \(\mu=10^{3.99}\):
comparison of time reparameterizations in the extreme stiffness regime, emphasis on boundary layer behaviour.}
\label{fig:vdp_case54b}
\end{figure}

To further clarify the extreme-stiffness behavior near fast transitions, Figure~\ref{fig:vdp_case54b} shows the evolution of individual components during a boundary layer. These plots suggest that the dominant error mode is a phase lead that manifests differently across methods. The proposed and extrema-based methods better preserve the qualitative transition structure, while the solver-directed method exhibits lower phase error but reduced fidelity in the transition back to the slow manifold, particularly for $y_2$. The proposed method is able to maintain a similar phase error to the solver-directed method while maintaining a fidelity to the slow-time mode present in the trajectory, transitioning between the two with greater accuracy than either other method.

\begin{table}[htbp]
\small
\centering
\caption{Van der Pol $\tau$-MSE (nondimensional) for all methods (On- vs Off-reference).}
\label{tab:vdp_tau_mse_nondim_4off}
\setlength{\tabcolsep}{6pt}
\renewcommand{\arraystretch}{1.15}
\begin{tabular}{llcccccc}
\toprule
& & \multicolumn{2}{c}{\textbf{Solver-Directed}} & \multicolumn{2}{c}{\textbf{Extrema-Based}} & \multicolumn{2}{c}{\textbf{Trajectory-Optimized}} \\
\cmidrule(lr){3-4}\cmidrule(lr){5-6}\cmidrule(lr){7-8}
\textbf{Block} & \textbf{Metric} & \textbf{ON} & \textbf{OFF} & \textbf{ON} & \textbf{OFF} & \textbf{ON} & \textbf{OFF} \\
\midrule
\multirow{5}{*}{\makecell[l]{$\tau$ MSE\\(nondim)}} 
& $y_1$  & $7.49\times10^{-2}$ & $9.14\times10^{-2}$ & $2.50\times10^{-3}$ & $2.24\times10^{-3}$ & $5.68\times10^{-3}$ & $6.23\times10^{-3}$ \\
& $y_2$  & $1.05\times10^{-1}$ & $1.21\times10^{-1}$ & $3.30\times10^{-3}$ & $2.98\times10^{-3}$ & $6.95\times10^{-3}$ & $7.70\times10^{-3}$ \\
& Time   & $4.68\times10^{-4}$ & $5.15\times10^{-4}$ & $3.20\times10^{-5}$ & $3.17\times10^{-5}$ & $3.31\times10^{-5}$ & $3.53\times10^{-5}$ \\
& State  & $9.00\times10^{-2}$ & $1.06\times10^{-1}$ & $2.90\times10^{-3}$ & $2.61\times10^{-3}$ & $6.32\times10^{-3}$ & $6.96\times10^{-3}$ \\
& All    & $6.02\times10^{-2}$ & $7.10\times10^{-2}$ & $1.95\times10^{-3}$ & $1.75\times10^{-3}$ & $4.22\times10^{-3}$ & $4.65\times10^{-3}$ \\
\bottomrule
\end{tabular}
\end{table}

\begin{table}[htbp]
\small
\centering
\caption{Van der Pol MSIE (dimensional) for all methods (On- vs Off-reference).}
\label{tab:vdp_msie_dim_4off}
\setlength{\tabcolsep}{6pt}
\renewcommand{\arraystretch}{1.15}
\begin{tabular}{llcccccc}
\toprule
& & \multicolumn{2}{c}{\textbf{Solver-Directed}} & \multicolumn{2}{c}{\textbf{Extrema-Based}} & \multicolumn{2}{c}{\textbf{Trajectory-Optimized}} \\
\cmidrule(lr){3-4}\cmidrule(lr){5-6}\cmidrule(lr){7-8}
\textbf{Block} & \textbf{Metric} & \textbf{ON} & \textbf{OFF} & \textbf{ON} & \textbf{OFF} & \textbf{ON} & \textbf{OFF} \\
\midrule
\multirow{3}{*}{\makecell[l]{MSIE\\(dim)}} 
& $y_1$  & $6.14\times10^{-1}$ & $5.17\times10^{-1}$ & $3.44\times10^{-1}$ & $3.82\times10^{-1}$ & $2.04\times10^{-1}$ & $2.21\times10^{-1}$ \\
& $y_2$  & $1.96\times10^{-3}$ & $1.27\times10^{-3}$ & $5.79\times10^{-4}$ & $5.41\times10^{-4}$ & $5.22\times10^{-4}$ & $1.78\times10^{-3}$ \\
& State  & $3.08\times10^{-1}$ & $2.59\times10^{-1}$ & $1.72\times10^{-1}$ & $1.91\times10^{-1}$ & $1.02\times10^{-1}$ & $1.11\times10^{-1}$ \\
\bottomrule
\end{tabular}
\end{table}

\paragraph{Discussion of problem two results.}
The van der Pol oscillator highlights the importance of time reparameterizations that accommodate both slow-manifold dynamics and fast transient layers while maintaining a regular and learnable clock. As shown in Table \ref{tab:vdp_tau_mse_nondim_4off}, all three approaches reduce stiffness sufficiently to permit stable explicit integration; however, substantial differences emerge in the quality and robustness of the resulting time maps. The solver-directed strategy exhibits significantly larger nondimensional $\tau$-MSE values—on the order of $10^{-1}$ for the state components, despite this, the method has the lowest phase error of the methods considered. In contrast, the extrema-based and proposed methods achieve reductions of more than an order of magnitude, with the proposed approach maintaining consistently low errors across all metrics and only modest degradation off-reference. Although the proposed method exhibits slightly higher $\tau$-MSE values than the extrema-based approach in Table \ref{tab:vdp_tau_mse_nondim_4off}, it consistently produces more accurate state trajectories in physical time, as reflected by the lower MSIE values in Table \ref{tab:vdp_msie_dim_4off}. This behavior underscores that $\tau$-MSE is not itself a measure of model fidelity, but rather a local metric of discrepancies in the learned time map. A trajectory with small pointwise errors may still allocate stretched-time resolution inefficiently, leading to degraded reconstruction of the physical dynamics. In contrast, the proposed method prioritizes the quality of the resulting trajectory by optimizing the time reparameterization to yield smoother and more uniformly traversed dynamics in stretched time. Consequently, even when the ML-ROM differs locally from the reference, the induced trajectory better captures the dominant dynamical features—particularly across the transitions present in this problem. This resulted in the lower observed error in physical time.

\subsection{Problem 3: HIRES Chemical Kinetics System}

The third benchmark is the HIRES system, a stiff nonlinear chemical kinetics model with
eight coupled ODEs,
\begin{equation}
\dot{\bm{y}}(t;\mu)=\bm{f}(\bm{y}(t;\mu)),\qquad
\bm{y}(0)=\bm{y}_0,\qquad
\bm{y}(t)\in\mathbb{R}^8,
\end{equation}
with dynamics
\begin{align}
\dot{y}_1 &= -1.71y_1 + 0.43y_2 + 8.32y_3 + 0.0007,\\
\dot{y}_2 &=  1.71y_1 - 8.75y_2,\\
\dot{y}_3 &= -10.03y_3 + 0.43y_4 + 0.035y_5,\\
\dot{y}_4 &=  8.32y_2 + 1.71y_3 - 1.12y_4,\\
\dot{y}_5 &= -1.745y_5 + 0.43y_6 + 0.43y_7,\\
\dot{y}_6 &= -\mu\,y_6 y_8 + 0.69y_4 + 1.71y_5 - 0.43y_6 + 0.69y_7,\\
\dot{y}_7 &=  \mu\,y_6 y_8 - 1.81y_7,\\
\dot{y}_8 &= -\mu\,y_6 y_8 + 1.81y_7.
\end{align}
The stiffness enters through both the disparity in chemical reaction rates as well as the nonlinear reaction term \(\mu y_6 y_8\), which most
directly affects \(y_6\), \(y_7\), and \(y_8\), while the remaining components are coupled
indirectly through the linear source terms.

Training data were generated for
\[
\mu \in \{10^{2},10^{2.1},\ldots,10^{3.9},10^{4}\},
\]
with off-reference testing at
\[
\mu_{\mathrm{test}} \in \{10^{2.025},10^{2.675},10^{3.325},10^{3.975}\}.
\]
A distinctive feature of HIRES is the presence of two separated stiff episodes: an early transient (first \(\mathcal{O}(10)\) seconds) in which the solution rapidly collapses onto a slow manifold, and a late-time event (around 350) seconds) in which depletion the of \(y_5\) and \(y_6\) precipitates rapid changes in \(y_7\) and \(y_8\). This strength of this second stiff event increases with $\mu$. Because the system contains eight state variables, only a representative subset is plotted to maintain clarity. The variables $y_1$, $y_5$, and $y_8$ were chosen to highlight distinct dynamical features: $y_1$ captures the dominant response during the initial stiff event, $y_5$ reflects slow-manifold evolution between boundary layers, and $y_8$ exhibits behavior characteristic of both stiff transients.

\begin{figure}[h!]
\centering
\begin{subfigure}{0.32\textwidth}
  \centering
  \includegraphics[width=\linewidth]{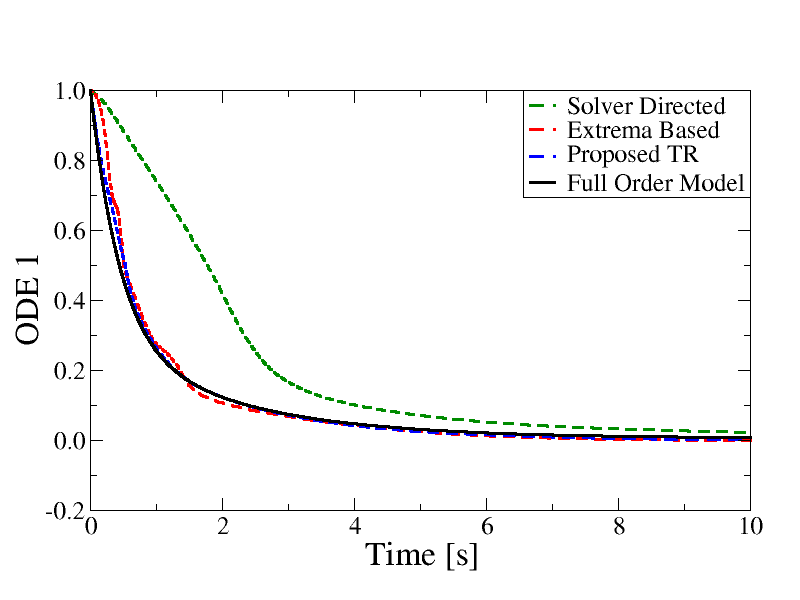}
  \caption{ODE one}
\end{subfigure}\hfill
\begin{subfigure}{0.32\textwidth}
  \centering
  \includegraphics[width=\linewidth]{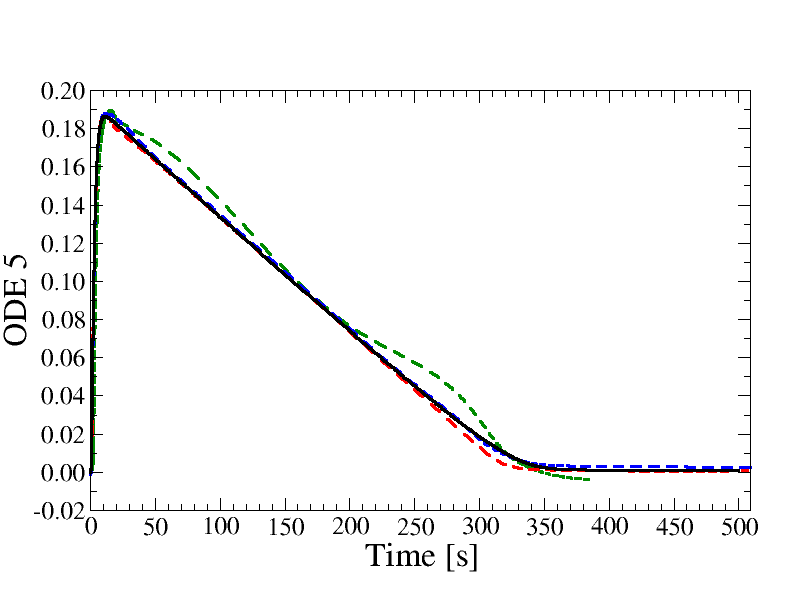}
  \caption{ODE five}
\end{subfigure}\hfill
\begin{subfigure}{0.32\textwidth}
  \centering
  \includegraphics[width=\linewidth]{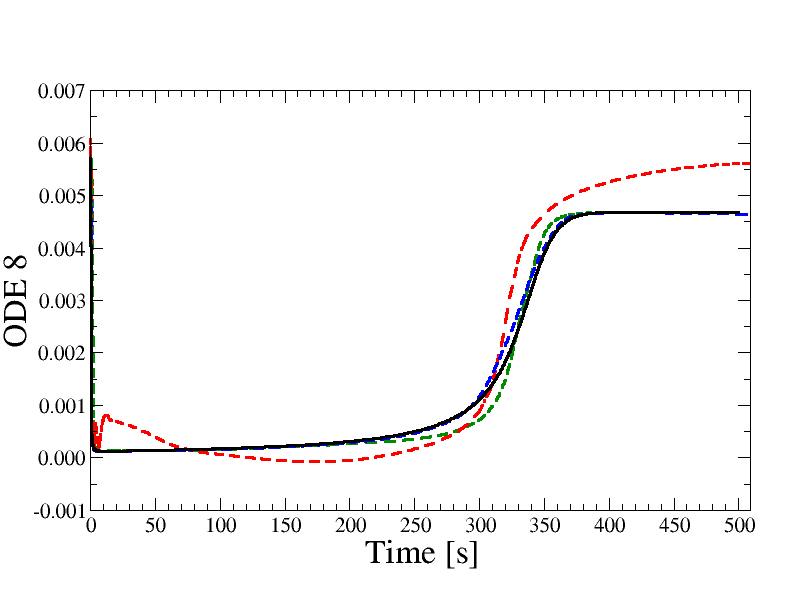}
  \caption{ODE eight}
\end{subfigure}
\caption{HIRES system, off-reference test case \(\mu=10^{2.025}\).}
\label{fig:hires_case2}
\end{figure}

\paragraph{Moderate stiffness regime.}
Figure~\ref{fig:hires_case2} shows the state trajectories for \(\mu=10^{2.025}\), a regime in which the HIRES system exhibits clear but still moderately separated stiffness scales. The late-time stiff event is particularly informative: \(y_7\) and \(y_8\) undergo rapid, monotone transitions over a short physical-time window. Capturing this behavior requires the inferred clock to allocate sufficient stretched-time resolution despite the absence of extrema. Because this late-time event is monotone, the extrema-based heuristic—which relies the presence of extrema near stiff events—fails to significantly slow down the clock in this region. As a result, fidelity is lost in \(y_7\) and \(y_8\), as reflected both visually and quantitatively.

\begin{figure}[h!]
\centering
\begin{subfigure}{0.32\textwidth}
  \centering
  \includegraphics[width=\linewidth]{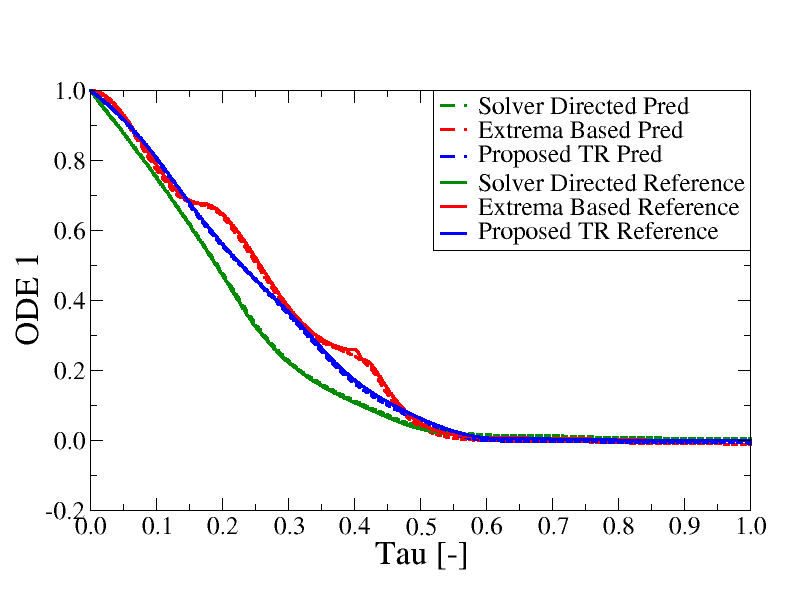}
  \caption{ODE one}
\end{subfigure}\hfill
\begin{subfigure}{0.32\textwidth}
  \centering
  \includegraphics[width=\linewidth]{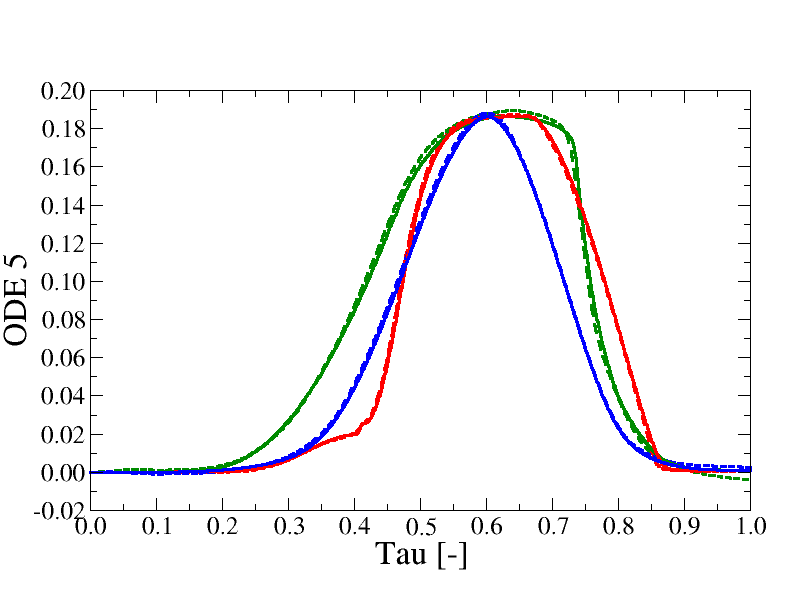}
  \caption{ODE five}
\end{subfigure}\hfill
\begin{subfigure}{0.32\textwidth}
  \centering
  \includegraphics[width=\linewidth]{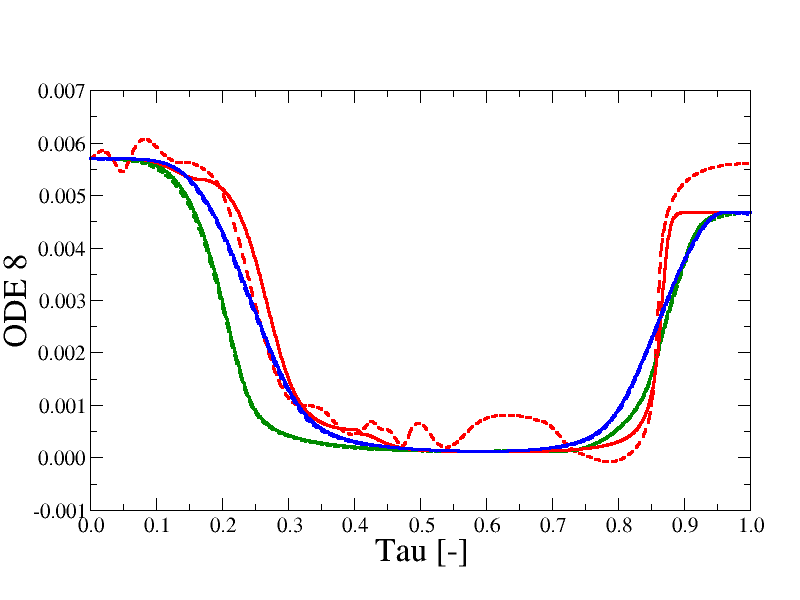}
  \caption{ODE eight}
\end{subfigure}
\caption{HIRES system, off-reference test case \(\mu=10^{2.025}\).}
\label{fig:hires_case2_tau}
\end{figure}

\begin{figure}[h!]
\centering
\begin{subfigure}{0.5\textwidth}
  \centering
  \includegraphics[width=\linewidth]{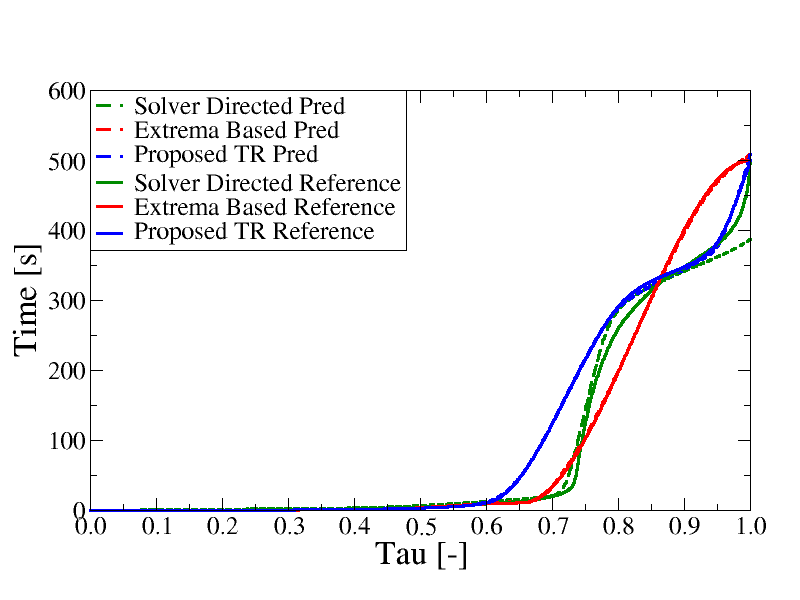}
  \caption{Time vs. $\tau$}
\end{subfigure}\hfill
\begin{subfigure}{0.5\textwidth}
  \centering
  \includegraphics[width=\linewidth]{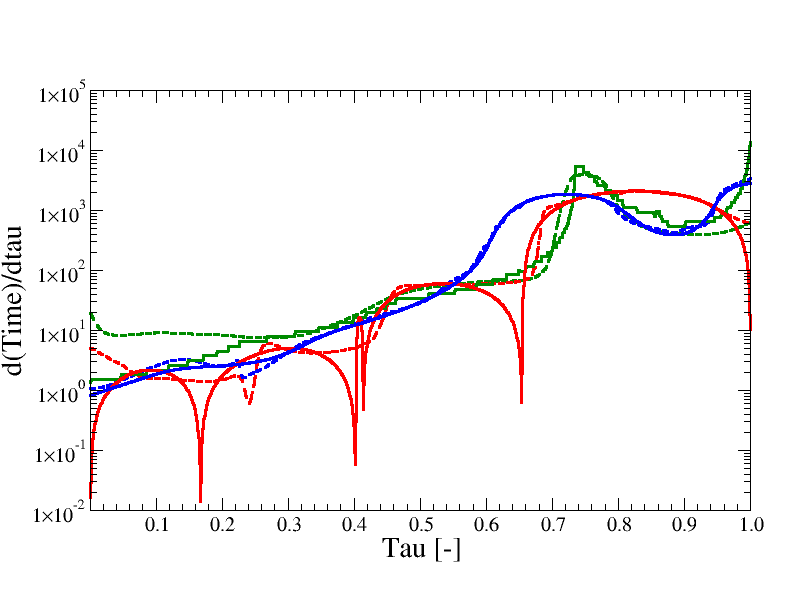}
  \caption{Derivative of time with respect to $\tau$}
\end{subfigure}\hfill
\caption{HIRES system, off-reference test case \(\mu=10^{2.025}\).}
\label{fig:hires_case2_time}
\end{figure}

The mechanism behind this behavior is evident in Figure~\ref{fig:hires_case2_tau}, where the second stiff event evolves much faster in \(\tau\) than in \(t\). The solver-directed method exhibits a different failure mode: although the reparameterized state trajectories are relatively smooth and therefore easy for the state network to approximate, the inferred time mapping cuts the simulation significantly short. The resulting physical-time trajectory, shown in Figure~\ref{fig:hires_case2_time}, contains regions of rapid acceleration that the ML-ROM is unable to reproduce accurately. By contrast, the proposed method produces smoother, more gradually varying time curves while still resolving the late-time stiff event, leading to trajectories that are consistently easier for the ML-ROM to learn and extrapolate.

\begin{figure}[h!]
\centering
\begin{subfigure}{0.32\textwidth}
  \centering
  \includegraphics[width=\linewidth]{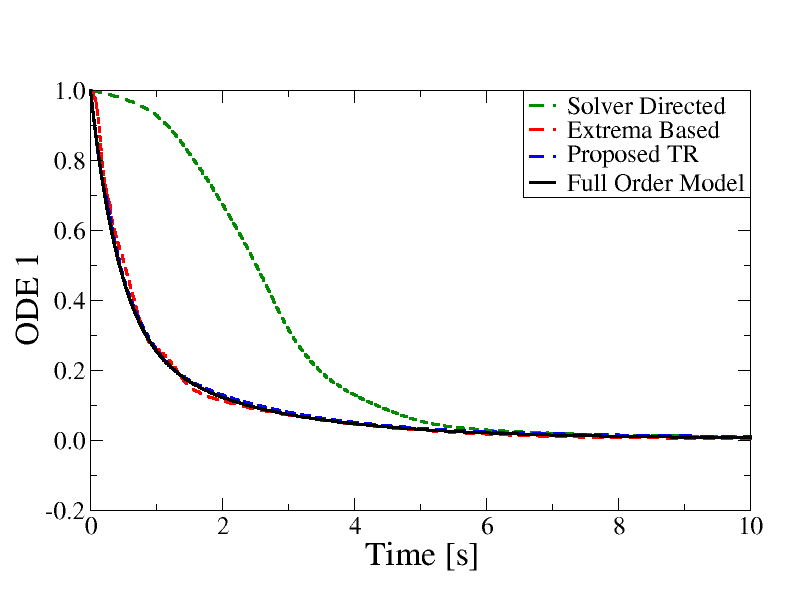}
  \caption{ODE one}
\end{subfigure}\hfill
\begin{subfigure}{0.32\textwidth}
  \centering
  \includegraphics[width=\linewidth]{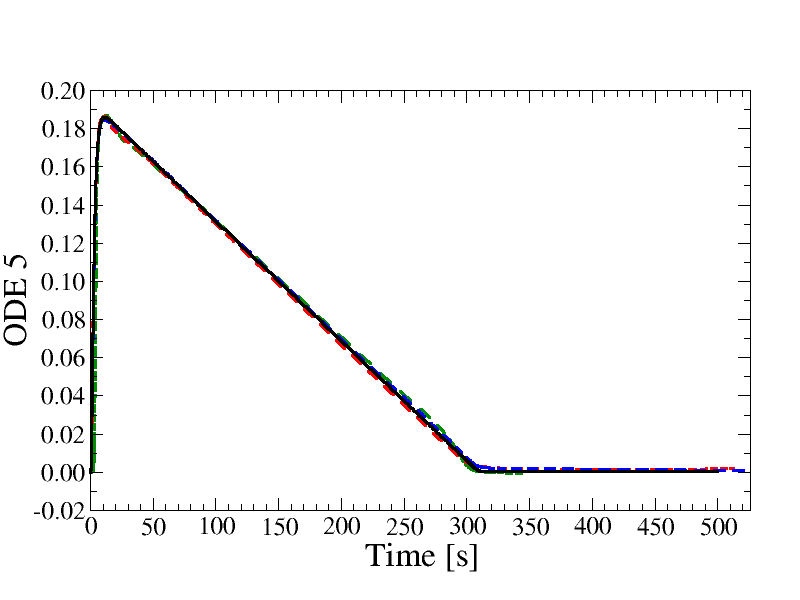}
  \caption{ODE five}
\end{subfigure}\hfill
\begin{subfigure}{0.32\textwidth}
  \centering
  \includegraphics[width=\linewidth]{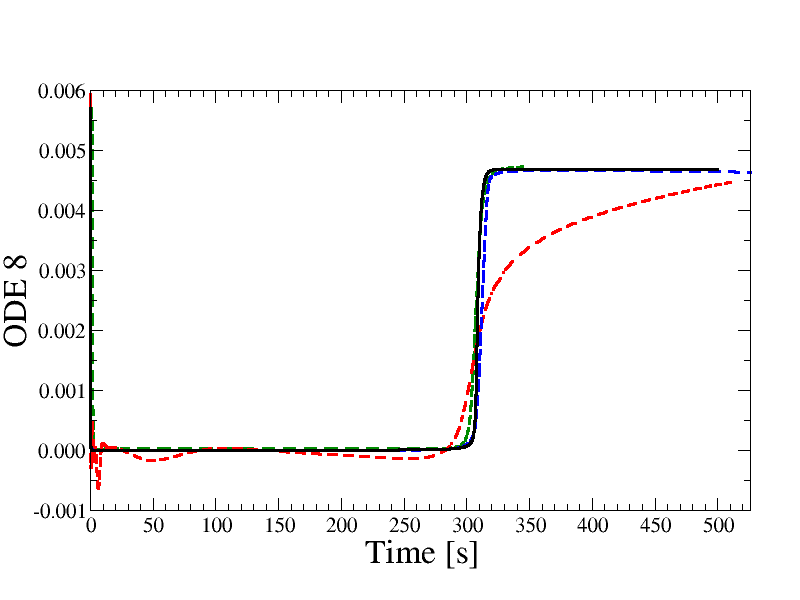}
  \caption{ODE eight}
\end{subfigure}
\caption{HIRES system, off-reference test case \(\mu=10^{3.975}\).}
\label{fig:hires_case23}
\end{figure}

\begin{figure}[h!]
\centering
\begin{subfigure}{0.32\textwidth}
  \centering
  \includegraphics[width=\linewidth]{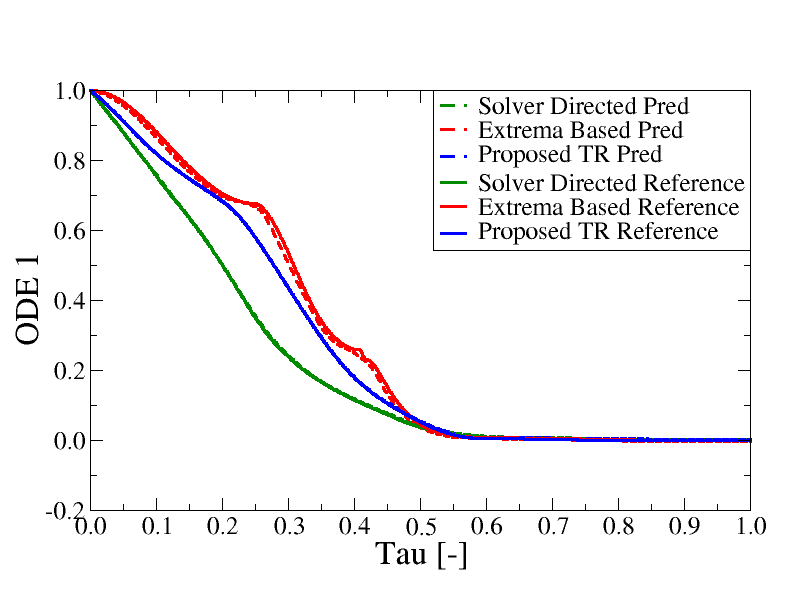}
  \caption{ODE one}
\end{subfigure}\hfill
\begin{subfigure}{0.32\textwidth}
  \centering
  \includegraphics[width=\linewidth]{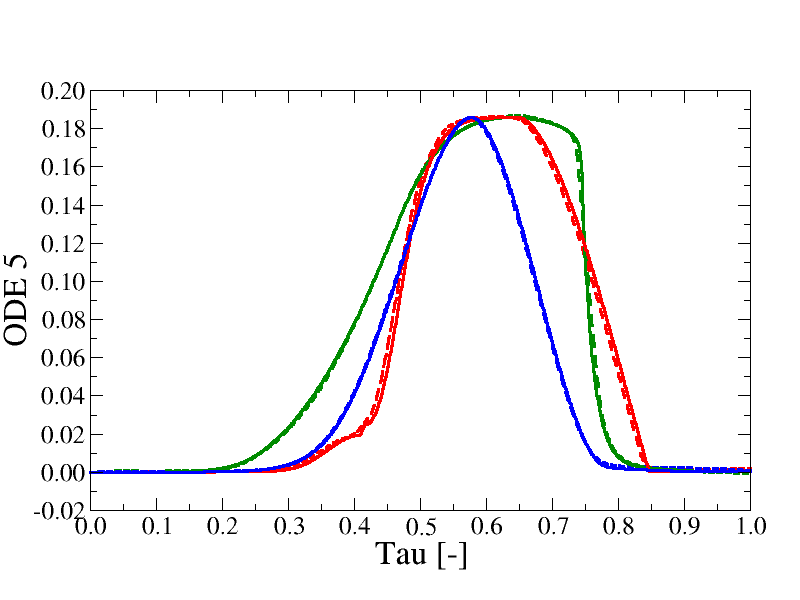}
  \caption{ODE five}
\end{subfigure}\hfill
\begin{subfigure}{0.32\textwidth}
  \centering
  \includegraphics[width=\linewidth]{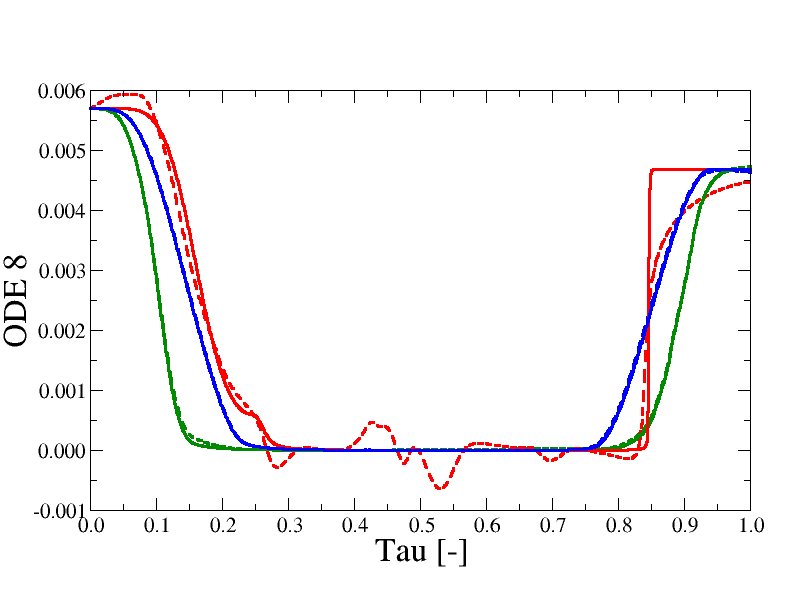}
  \caption{ODE eight}
\end{subfigure}
\caption{HIRES system, off-reference test case \(\mu=10^{3.975}\).}
\label{fig:hires_case23_tau}
\end{figure}

\paragraph{Extreme stiffness regime.}
Figure~\ref{fig:hires_case23} presents the corresponding results for \(\mu=10^{3.975}\), where both stiffness mechanisms—the initial collapse onto the slow manifold and the late-time event—become more abrupt in physical time. At this level of stiffness, differences in how the methods allocate stretched-time resolution are strongly amplified. The solver-directed method again struggles to capture the early-time transient and fails to reach the end of the simulation horizon, despite producing relatively smooth state curves. The extrema-based method continues to inadequately resolve the second stiff event; as seen in the reparameterized trajectories in Figure~\ref{fig:hires_case23_tau}, the late-time dynamics are effectively compressed, producing a challenging target for the ML-ROM.

\begin{figure}[h!]
\centering
\begin{subfigure}{0.5\textwidth}
  \centering
  \includegraphics[width=\linewidth]{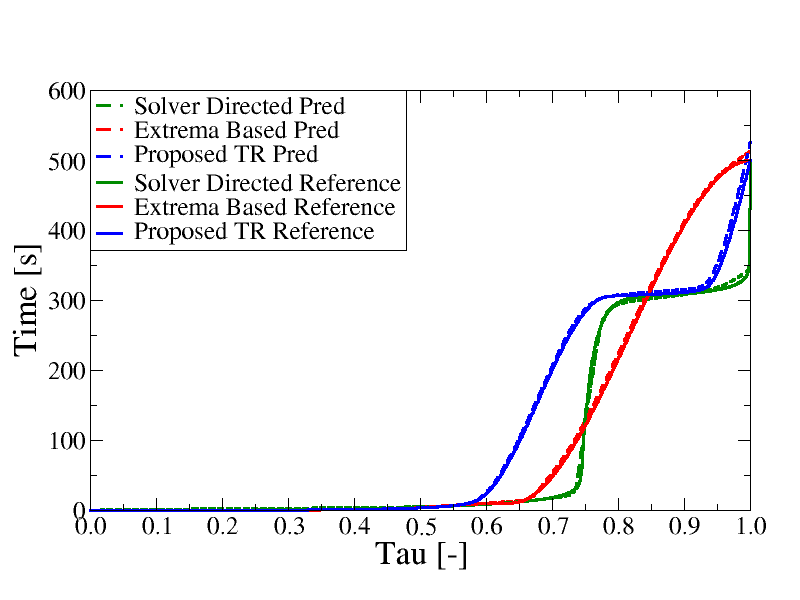}
  \caption{Time vs. $\tau$}
\end{subfigure}\hfill
\begin{subfigure}{0.5\textwidth}
  \centering
  \includegraphics[width=\linewidth]{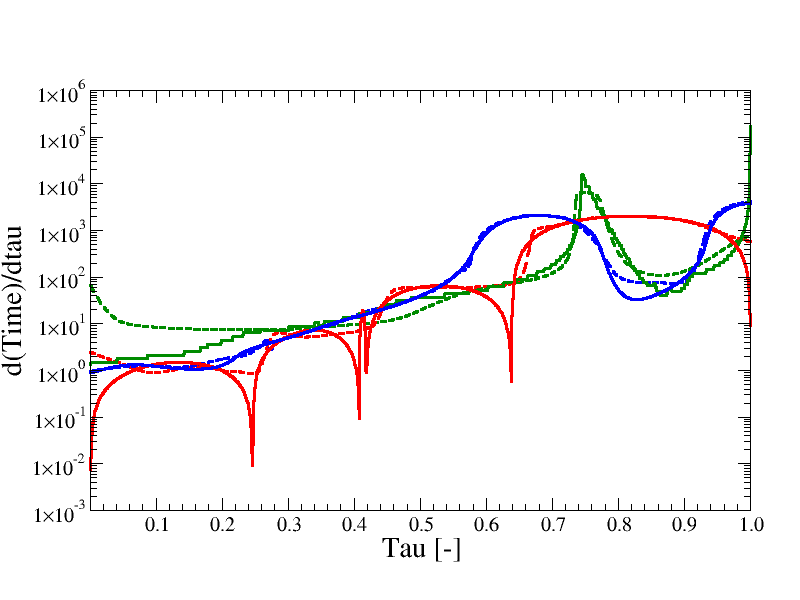}
  \caption{Derivative of time with respect to $\tau$}
\end{subfigure}\hfill
\caption{HIRES system, off-reference test case \(\mu=10^{3.975}\).}
\label{fig:hires_case23_time}
\end{figure}

The time mappings in Figure~\ref{fig:hires_case23_time} further clarify these issues. The solver-directed reference trajectory undergoes an extreme acceleration—over two-order-of-magnitude increase in time velocity over the final one percent of the \(\tau\)-domain—rendering it essentially unlearnable by the ML-ROM. Meanwhile, the extrema-based time curve exhibits rapid oscillations in its derivative, which again degrade the accuracy of the learned model. The proposed method avoids both issues, maintaining smooth derivatives and transitions even in this extreme stiffness regime.

\paragraph{Discussion of problem three results.}

The HIRES benchmark combines nonlinear reaction kinetics, strong parameter dependence, and two well-separated stiff episodes, making it a stringent test for learning-compatible time reparameterizations. The quantitative results in Tables~\ref{tab:hires_tau_mse_nondim} and~\ref{tab:hires_msie_dim} reinforce the qualitative observations from the trajectory plots and that the proposed method consistently outperforms the baselines.

Table~\ref{tab:hires_tau_mse_nondim} reports nondimensional \(\tau\)-MSE values for individual state components, the inferred time map, and aggregate metrics. Across all variables, the proposed method achieves errors that are typically one to two orders of magnitude smaller than those of both the solver-directed and the extrema-based, in both on- and off-reference settings. This improvement is especially pronounced for \(y_7\) and \(y_8\), which are directly affected by the stiff nonlinear reaction term. For these components, the extrema-based method exhibits \(\tau\)-MSE values on the order of \(10^{-3}\), whereas the proposed method maintains errors near \(10^{-5}\), indicating a far more faithful allocation of stretched-time resolution during the late-time stiff event. The reduced time-map error further demonstrates that the proposed approach produces a smoother and more accurate clock, which directly impacts the learnability of the downstream ML-ROM.

The dimensional MSIE results in Table~\ref{tab:hires_msie_dim} provide complementary insight. While the solver-directed method yields large state errors for variables such as \(y_1\), \(y_4\), and \(y_6\)—reflecting its difficulty with early-time stiffness and truncated simulations—the proposed method consistently achieves the lowest state-level MSIE across nearly all components. Importantly, these improvements persist in the off-reference setting, demonstrating robustness. The state-aggregated MSIE is reduced by nearly two orders of magnitude relative to the solver-directed results and remains competitive or superior to the extrema-based method, even for variables indirectly affected by stiffness.

Taken together, these results highlight a key distinction between methods that merely produce smooth reparameterized state trajectories and those that generate learning-compatible clocks. Irregular or overly aggressive time reparameterizations can produce sharp accelerations in the reparameterized trajectory or in the clock that introduce large gradient magnitudes during training and hinder generalization across \(\mu\). By instead optimizing for smoothness in the reparameterized trajectory, the proposed method aligns the time map with the representational capacity of the ML-ROM. This alignment is crucial for complex multiscale systems like HIRES, where accurate state prediction depends heavily on the structure of the learned clock.

\begin{table}[htbp]
\small
\centering
\caption{HIRES $\tau$-MSE (nondimensional) for all methods (ON- vs OFF-reference).}
\label{tab:hires_tau_mse_nondim}
\setlength{\tabcolsep}{6pt}
\renewcommand{\arraystretch}{1.15}
\begin{tabular}{llcccccc}
\toprule
& & \multicolumn{2}{c}{\textbf{Solver-Directed}} & \multicolumn{2}{c}{\textbf{Extrema-Based}} & \multicolumn{2}{c}{\textbf{Trajectory-Optimized}} \\
\cmidrule(lr){3-4}\cmidrule(lr){5-6}\cmidrule(lr){7-8}
\textbf{Block} & \textbf{Metric} & \textbf{ON} & \textbf{OFF} & \textbf{ON} & \textbf{OFF} & \textbf{ON} & \textbf{OFF} \\
\midrule
\multirow{11}{*}{\makecell[l]{$\tau$ MSE\\(nondim)}}
& $y_1$  & $1.65\times10^{-5}$ & $1.68\times10^{-5}$ & $6.86\times10^{-5}$ & $7.94\times10^{-5}$ & $1.11\times10^{-5}$ & $1.93\times10^{-5}$ \\
& $y_2$  & $1.00\times10^{-3}$ & $9.87\times10^{-4}$ & $3.30\times10^{-4}$ & $3.79\times10^{-4}$ & $4.52\times10^{-5}$ & $4.83\times10^{-5}$ \\
& $y_3$  & $1.49\times10^{-3}$ & $1.51\times10^{-3}$ & $4.31\times10^{-4}$ & $4.49\times10^{-4}$ & $9.62\times10^{-5}$ & $1.17\times10^{-4}$ \\
& $y_4$  & $2.18\times10^{-4}$ & $2.10\times10^{-4}$ & $2.12\times10^{-4}$ & $2.32\times10^{-4}$ & $4.81\times10^{-5}$ & $4.99\times10^{-5}$ \\
& $y_5$  & $1.43\times10^{-4}$ & $1.84\times10^{-4}$ & $2.95\times10^{-4}$ & $2.18\times10^{-4}$ & $3.70\times10^{-5}$ & $4.29\times10^{-5}$ \\
& $y_6$  & $2.02\times10^{-4}$ & $2.51\times10^{-4}$ & $2.66\times10^{-4}$ & $1.94\times10^{-4}$ & $3.30\times10^{-5}$ & $3.63\times10^{-5}$ \\
& $y_7$  & $2.56\times10^{-5}$ & $3.06\times10^{-5}$ & $4.43\times10^{-3}$ & $4.73\times10^{-3}$ & $3.48\times10^{-5}$ & $2.76\times10^{-5}$ \\
& $y_8$  & $2.36\times10^{-5}$ & $2.92\times10^{-5}$ & $4.58\times10^{-3}$ & $4.71\times10^{-3}$ & $3.46\times10^{-5}$ & $2.79\times10^{-5}$ \\
& Time   & $2.65\times10^{-4}$ & $3.52\times10^{-4}$ & $2.79\times10^{-4}$ & $4.92\times10^{-5}$ & $3.18\times10^{-5}$ & $3.46\times10^{-5}$ \\
& State  & $3.91\times10^{-4}$ & $4.03\times10^{-4}$ & $1.33\times10^{-3}$ & $1.37\times10^{-3}$ & $4.25\times10^{-5}$ & $4.62\times10^{-5}$ \\
& All    & $3.77\times10^{-4}$ & $3.97\times10^{-4}$ & $1.21\times10^{-3}$ & $1.23\times10^{-3}$ & $4.13\times10^{-5}$ & $4.49\times10^{-5}$ \\
\bottomrule
\end{tabular}
\end{table}

\begin{table}[htbp]
\small
\centering
\caption{HIRES MSIE (dimensional) for all methods (ON- vs OFF-reference).}
\label{tab:hires_msie_dim}
\setlength{\tabcolsep}{6pt}
\renewcommand{\arraystretch}{1.15}
\begin{tabular}{llcccccc}
\toprule
& & \multicolumn{2}{c}{\textbf{Solver-Directed}} & \multicolumn{2}{c}{\textbf{Extrema-Based}} & \multicolumn{2}{c}{\textbf{Trajectory-Optimized}} \\
\cmidrule(lr){3-4}\cmidrule(lr){5-6}\cmidrule(lr){7-8}
\textbf{Block} & \textbf{Metric} & \textbf{ON} & \textbf{OFF} & \textbf{ON} & \textbf{OFF} & \textbf{ON} & \textbf{OFF} \\
\midrule
\multirow{9}{*}{\makecell[l]{MSIE\\(dim)}}
& $y_1$  & $1.55\times10^{-3}$ & $1.58\times10^{-3}$ & $2.47\times10^{-5}$ & $3.55\times10^{-5}$ & $1.70\times10^{-5}$ & $2.55\times10^{-5}$ \\
& $y_2$  & $5.54\times10^{-5}$ & $5.54\times10^{-5}$ & $1.80\times10^{-6}$ & $2.38\times10^{-6}$ & $9.74\times10^{-7}$ & $9.43\times10^{-7}$ \\
& $y_3$  & $3.76\times10^{-6}$ & $3.78\times10^{-6}$ & $1.88\times10^{-7}$ & $1.85\times10^{-7}$ & $1.34\times10^{-8}$ & $1.53\times10^{-8}$ \\
& $y_4$  & $7.42\times10^{-4}$ & $7.55\times10^{-4}$ & $1.14\times10^{-5}$ & $1.45\times10^{-5}$ & $1.11\times10^{-5}$ & $1.25\times10^{-5}$ \\
& $y_5$  & $7.69\times10^{-5}$ & $7.89\times10^{-5}$ & $1.99\times10^{-6}$ & $2.56\times10^{-6}$ & $1.02\times10^{-6}$ & $1.29\times10^{-6}$ \\
& $y_6$  & $1.17\times10^{-3}$ & $1.20\times10^{-3}$ & $2.04\times10^{-5}$ & $2.73\times10^{-5}$ & $8.05\times10^{-6}$ & $9.98\times10^{-6}$ \\
& $y_7$  & $1.92\times10^{-7}$ & $1.81\times10^{-7}$ & $3.15\times10^{-7}$ & $3.33\times10^{-7}$ & $1.24\times10^{-8}$ & $1.54\times10^{-8}$ \\
& $y_8$  & $1.93\times10^{-7}$ & $1.81\times10^{-7}$ & $2.57\times10^{-7}$ & $2.90\times10^{-7}$ & $1.24\times10^{-8}$ & $1.54\times10^{-8}$ \\
& State  & $4.49\times10^{-4}$ & $4.60\times10^{-4}$ & $7.62\times10^{-6}$ & $1.04\times10^{-5}$ & $4.77\times10^{-6}$ & $6.29\times10^{-6}$ \\
\bottomrule
\end{tabular}
\end{table}

\section{Discussion}
\label{sec:discussion}

This work studies time reparameterization (TR) as a mechanism for mitigating stiffness in neural ODE--based reduced-order models. The central empirical observation is that the usefulness of a time map is governed by both how much it ``slows down'' stiff regions and by whether it produces a clock and associated state trajectories that are learnable and smooth across parameters. In particular, clocks with sharp accelerations or oscillatory derivatives can reintroduce stiffness-like behavior even when the reparameterized state trajectory appears smooth. The trajectory-optimized method proposed here is designed to address this failure mode directly: rather than inferring a time map from solver behavior or from a heuristic, it computes a speed profile that minimizes an acceleration-based functional along the trajectory, yielding time maps whose derivatives are smoother and whose parametric dependence is more regular.

Figures~\ref{fig:tol3} and~\ref{fig:tol6} demonstrate that solver-directed time reparameterizations are highly sensitive to integration tolerances, producing inconsistent stretched-time representations that can depend on numerical settings rather than the underlying dynamics. In contrast, the trajectory-optimized and extrema-based approaches yield more stationary and repeatable clocks, improving robustness to noise.  Notably, although the proposed method targets smoothness in stretched time $\tau$, it also produces smooth dependence on $\mu$, which is beneficial for neural ODE training by promoting coherent trajectories accross parameters.

Two complementary evaluation metrics are used throughout: (i) a nondimensional $\tau$-MSE, which measures prediction error in the stretched-time training coordinate, including the learned time map, and (ii) a reparameterization-invariant MSIE in physical time, which measures fidelity to the true trajectory regardless of the chosen clock. Together, these quantify a key distinction: a time reparameterization can be locally accurate in $\tau$ while still producing a poor or difficult clock for reconstructing dynamics in physical time. This effect is most visible in the Van der Pol benchmark, where the extrema-based method attains smaller $\tau$-MSE than the proposed approach for some metrics (Table~\ref{tab:vdp_tau_mse_nondim_4off}), yet the proposed approach yields lower physical-time error in the aggregated MSIE (Table~\ref{tab:vdp_msie_dim_4off}). This motivates reporting both: $\tau$-MSE measures learnability in stretched time, while MSIE evaluates accuracy in physical time.

The SLS benchmark isolates stiffness without nonlinear geometric complexity and therefore highlights how reparameterizations interact with parametric variation. In the extreme stiffness regime, the proposed method produces the lowest errors in both the time map and the state in $\tau$ (Table~\ref{tab:sls_tau_mse_nondim}), with especially large gains in the learned time component, and achieves the lowest aggregated physical-time error (Table~\ref{tab:sls_msie_dim}). The solver-directed reparameterization may introduce cusp-like transitions in $\tau$, while the extrema-based heuristic can yield highly oscillatory time derivatives. These features make $\alpha$ harder to approximate and produce large variations in the gradient of the reparameterized vector field. The SLS results therefore support the premise that clock smoothness drives learnability, particularly when small changes in $\mu$ should not induce large changes in the time map.

The van der Pol results demonstrate that time maps must balance phase accuracy with smooth resolution of fast transitions. The solver-directed approach produces comparatively large $\tau$-MSE for the state  (Table~\ref{tab:vdp_tau_mse_nondim_4off}), yet often displays low phase error in physical time. The arc-length-based methods, by concentrating resolution near high-variation segments, better preserve transition shapes but induced more phase lead. The TOTR approach is the best method in MSIE (Table~\ref{tab:vdp_msie_dim_4off}) and exhibits consistent behavior off-reference. This case emphasizes that a ``good'' clock must both be learnable in stretched time and lead to accurate reconstruction of the physical-time dynamics.

HIRES is the most stringent test because it features two separated stiff episodes: an early collapse to a slow manifold and a late-time depletion-driven event whose strength grows with $\mu$. The late-time stiff event in $y_7$ and $y_8$ is monotone and therefore does not affect the time dilation function of the extrema-based construction. The effect of this is clear, as the extrema-based  method exhibits $\tau$-MSE on the order of $10^{-3}$ for $y_7$ and $y_8$ (Table~\ref{tab:hires_tau_mse_nondim}), whereas the TOTR method remains near $10^{-5}$, indicating that it allocates stretched-time resolution effectively in the components affected by the nonlinear reaction term. Moreover, the approach achieves the smallest aggregated state MSIE (Table~\ref{tab:hires_msie_dim}) and does so robustly off-reference. By contrast, the solver-directed approach suffers from a premature termination due to a failure to capture the extreme acceleration in the time map near the end of the reference trajectory. This leads to a learned clock that fails to capture this aspect of the reference trajectory.

The proposed TOTR formulation differs from prior methods in that it treats TR as an optimization problem defined on a fixed geometric curve. In arc-length coordinates, the objective \eqref{eq:Js_final} penalizes both (i) rapid variation of traversal speed (tangential acceleration) and (ii) curvature-induced acceleration (normal component), subject to a fixed stretched-time budget. This construction targets the features that degrade explicit integrator stability and learning dynamics in $\tau$: large local accelerations correspond to rapidly changing derivatives that produce sharp gradients in the ML-ROM training procedure. By optimizing these quantities explicitly, the resulting time maps have smoother $\alpha(\tau)$ profiles and avoid the three major failure modes observed in the baselines: nonuniformity that can create cusps or late-time blow-up in $\mathrm{d}t/\mathrm{d}\tau$, rapid variations in the time map, and failure to capture stiff events that are not associated with extrema.

The experiments suggest the following practical guidance for TR in ML-ROM pipelines: (i) TR is most beneficial when training relies on explicit integration and uniform sampling, (ii) clocks should be evaluated not only for their ability to reduce stiffness but also for their learnability, smoothness, and robustness across parameter variation, and (iii) models should be selected using reparameterization-invariant physical-time metrics such as MSIE, since low $\tau$-MSE alone does not guarantee accurate recovery of physical trajectories. The results indicate that time map smoothness is crucial for ML-ROM learnability: smooth clocks reduce gradient magnitudes, improve optimization stability, and enhance generalization across parameters.

Several limitations of the present study should be noted. First, for the extrema-based baseline only the cubic-spline construction was considered. For the problems considered in this work, there was not a significant difference observed between the reparameterized trajectories of the two spline orders. The quintic-spline variant proposed in the same framework could potentially improve learnability in regimes where the cubic construction produces a difficult time map. A systematic comparison between spline orders is an important direction for future work. 

Second, the stretched-time horizon and sampling density in $\tau$ were fixed for all cases considered. In practice, a tradeoff exists between the number of uniform $\tau$ steps, which directly influences training and inference cost, and the achievable learning accuracy. Developing principled guidelines for selecting $\tau_f$ and the discretization level is beyond the scope of the present work. Fixing the number of $\tau$ steps enables a controlled and fair comparison between methods, but does not provide a comprehensive assessment of performance. Evaluating how different reparameterization strategies behave across varying $\tau$ resolutions therefore remains an important direction for future study.

Finally, all reported results were drawn directly from large, controlled hyperparameter searches in which the same sampling procedure and budget are applied to each method. This ensures fairness but does not represent the best achievable performance for any single method: in all cases, additional gains are almost certainly possible through manual, case-specific tuning of the training procedure. Consequently, the results should be interpreted as comparative performance under a fixed training regimen, rather than as absolute best-case accuracies for each approach.

Overall, the results support three main conclusions. First, all three time reparameterization algorithms are effective for enabling explicit, uniformly sampled neural ODE training in stiff regimes. Second, the quality of a TR strategy is not solely governed by its stiffness-reduction capability, but also by the regularity and learnability of the induced clock. Third, the TOTR method provides a robust and accurate alternative to the TR methods in the literature, yielding consistently lower physical-time error across all problems considered.

\section{Conclusion}
\label{sec:conclusion}

This work investigated time reparameterization (TR) as a mechanism for mitigating stiffness in neural ODE-based reduced-order models. While prior approaches have demonstrated that appropriate changes of the independent variable can reduce effective stiffness and enable explicit integration, their impact on learnability and parametric robustness has remained unclear. By systematically comparing different time reparameterizations across multiple stiff benchmark problems, this study clarifies the role of the time map itself as a central component of the learning problem.

The main contribution of this paper is the formulation of TOTR: a trajectory optimization TR problem posed in arc-length coordinates. By optimizing a traversal-speed that penalizes acceleration in stretched time, the proposed method constructs clocks and reparameterized trajectories that are smoother, better conditioned, and more consistent across parameter variations than those produced by existing methods. Importantly, this formulation targets the dynamics actually integrated during neural ODE training, rather than relying on indirect proxies for stiffness such as solver step sizes or the presence of local extrema. 

Across all three benchmark systems—a parameterized stiff linear system, the Van der Pol oscillator, and the HIRES chemical kinetics model—the proposed method consistently produced time maps that were easier to learn and that yielded lower physical-time prediction error under identical training regimens. Quantitative results typically demonstrated reductions of one to two orders of magnitude in stretched-time prediction error and substantial improvements in the reparameterization-invariant MSIE, particularly in regimes featuring severe stiffness. These gains were most pronounced in problems with multiple stiff episodes or monotone stiff events, where the baseline methods struggle to allocate resolution appropriately.

More broadly, the results highlight that effective stiffness mitigation in learning-based ROMs is not achieved solely by slowing down fast transients, but by constructing clocks that are smooth and robust to parametric variation. Irregular time maps can exhibit cusps, large variation in derivatives, or extreme accelerations, reintroducing stiffness-like behavior into the learning problem even when the reparameterized state trajectory appears smooth. By explicitly optimizing for smoothness in the transformed dynamics, the proposed approach aligns the time reparameterization with the representational and optimization properties of neural ODE models.

Taken together, these findings suggest that time reparameterization is a powerful tool in ML-ROM construction for stiff systems. The time-optimized time reparameterization method provides a principled and flexible framework for constructing clocks that are both numerically stable and learnable by neural ODE models. This offers a promising path toward robust, explicit, and scalable reduced-order modeling of multiscale dynamical systems.

\section*{Acknowledgments}
This work was supported through the Triad National Security by the Department of Energy - National Nuclear Security Administration program under federal award number 89233218CNA000001, entitled ``The Role of Second Law of Thermodynamics in Deriving Turbulence Models and Simplified Reaction Kinetics''. DL acknowledges support from the Laboratory Directed Research and Development - Exploratory Research at Los Alamos National Laboratory, under project number 20220318ER.

\bibliographystyle{plain}
\bibliography{sample} 

\end{document}